%% file: main.tex
\documentclass[letterpaper]{article} %
\usepackage[]{aaai25}  %
\usepackage{times}  %
\usepackage{helvet}  %
\usepackage{courier}  %
\usepackage[hyphens]{url}  %
\usepackage{graphicx} %
\usepackage{natbib}  %
\usepackage{caption} %
\usepackage{algorithm}
\usepackage{algorithmic}

\usepackage{newfloat}
\usepackage{listings}
\usepackage{graphicx} 
\usepackage{xcolor}

\DeclareCaptionStyle{ruled}{labelfont=normalfont,labelsep=colon,strut=off} %
\usepackage{listings}
\usepackage[utf8]{inputenc} %
\usepackage{booktabs}       %
\usepackage{amsfonts}       %
\usepackage{nicefrac}       %
\usepackage{microtype}      %
\usepackage{xcolor}         %
\usepackage{subcaption}

\newcommand{\atomic}{atomic\ }

\input{macros}

\newcommand{\appl}{App}
\newcommand{\prog}{Prog}
\newcommand{\reach}{Reach}
\newcommand{\areach}{AReach}
\newcommand{\just}{Just}
\newcommand{\landmark}{Land}
\newcommand{\validation}{Val}

\newcommand{\oone}{OpenAI\ o1} 
\newcommand{\gpt}{GPT-4o} 
\newcommand{\gptmini}{GPT-4o Mini}

\newcommand{\best}[1]{\boldsymbol{#1}}
\newcommand{\secondbest}[1]{\underline{#1}}
\newcommand{\bestofmini}[1]{^{\ast}#1}
\newcommand{\numofmodels}[0]{$22$}
\newcommand{\numofdomains}[0]{$13$}
\newcommand{\numoftasks}[0]{$7$}
\newcommand{\numoftraining}[0]{$8$}
\newcommand{\numoftesting}[0]{$5$}
\usepackage{amsmath, bm}
\usepackage{multirow}
\usepackage{tabularray}
\UseTblrLibrary{booktabs}

\title{ACPBench: Reasoning about Action, Change, and Planning}
\author{
}

\author{
Harsha Kokel,
Michael Katz, 
Kavitha Srinivas,
Shirin Sohrabi
}
\affiliations {
IBM Research
}

\begin{document}

\maketitle
\input{abstract}
\input{introduction}

\input{background_relatedwork}
\input{ACPBench}

\input{experiments}
\input{o1_experiments}

\section{Discussion and Future Work}

In this work, we introduce ACPBench---a collection of datasets to evaluate the ability of LLMs to reason about action, change and planning. By evaluating \numofmodels{} state-of-the-art LLMs of varying size, we find
these models underperform, even the largest ones,
especially on tasks such as plan validation and action reachability. On the other hand, we show that finetuning a small language model, Granite 8B, can improve its reasoning ability to bring it on par with the best performing models. Further, we observe that the fine-tuned model exhibits remarkable generalization ability to unseen domains in ACPBench as well as to a different task in PlanBench. Further, our investigation with \oone{} reasoning model indicates that OpenAI's multi-turn approach yields improvements for multi-choice questions but fails to make an impact on boolean questions in ACPBench.

Performance of LLMs is known to be sensitive to prompt text as well as prompt style. Hence, it is possible to elicit better performance from each of these models with prompt engineering.
In our work we do not modify prompts across models -- our objective in the evaluation is to set a baseline. 
We hope our benchmark serves as a useful resource for improving LLM abilities. We encourage creative solutions (not limited to prompt engineering) to improve LLM performance across various tasks of ACPBench.

\bibliography{abbrv-short,literatur,crossref-short}

\clearpage

\appendix
\input{appendix}

\end{document}

%% file: macros.tex
\def\defemb#1#2{\expandafter\def\csname #1\endcsname
                              {\relax\ifmmode #2\else\hbox{$#2$}\fi}}

\defemb{bara}{{\bar{a}}}
\defemb{barb}{{\bar{b}}}
\defemb{barc}{{\bar{c}}}
\defemb{barf}{{\bar{f}}}
\defemb{barv}{{\bar{v}}}

{\hfill $\square$ \par \addvspace{\bigskipamount}}

\newcommand{\pre}{\ensuremath{\mathit{pre}}}

\newcommand{\add}{\ensuremath{\mathit{add}}}
\newcommand{\del}{\ensuremath{\mathit{del}}}

\newcommand{\actions}{O}

\newcommand{\ucausalG}[1]{\mbox{\sl G}_{CG}}

%% file: abstract.tex
\begin{abstract}

There is an increasing body of work using Large Language Models (LLMs) as agents for orchestrating workflows and making decisions in domains that require planning and multi-step reasoning. 
As a result, it is imperative to evaluate LLMs on core skills required for planning. In this work, we present ACPBench, a  benchmark for evaluating the reasoning tasks in the field of planning. The benchmark consists of $7$ reasoning tasks over $13$ planning domains. 
The collection is constructed from 
planning domains described in a formal language. 
This allows us to synthesize problems with provably correct solutions across many tasks and domains. Further, it allows us the luxury of scale without additional human effort, i.e., many additional problems can be created automatically. 
Our
extensive evaluation of \numofmodels{} 
LLMs and OpenAI o1 reasoning models highlight the significant gap in the reasoning capability
of the LLMs. 
Our findings with \oone{}, a multi-turn reasoning model, reveal significant gains in performance on multiple-choice questions, yet surprisingly, no notable progress is made on boolean questions.

ACPBench collection is available at the following link: \url{https://ibm.github.io/ACPBench}
\end{abstract}

%% file: introduction.tex
\section{Introduction}

Recent research has explored the potential of using Large Language Models (LLMs) as reasoners for solving multi-step reasoning problems~\cite{chu-etal-2024-cotsurvey}. Building on their success in certain reasoning tasks and benchmarks, there is a growing interest in using LLMs as agents for orchestrating workflows and making decisions in domains that require planning~\cite{Huang_survey,WangMFZYZCTCLZWW24_llmagent}. This is a promising area of research, with potential applications in various fields. However, there is a lack of systematic evaluation of LLMs reasoning and planning capabilities.

\begin{figure}[!t]
    \centering
\includegraphics[width=\columnwidth]{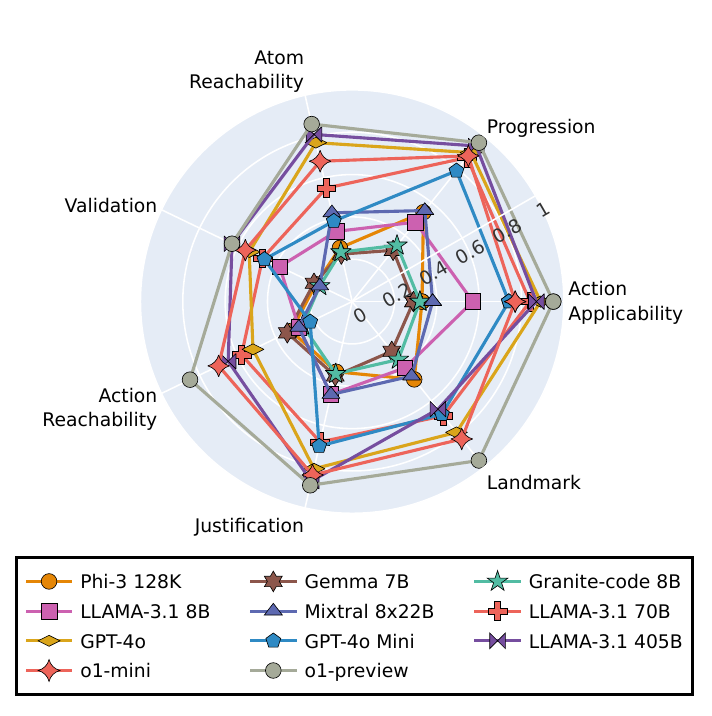}
    \caption{\footnotesize Performance of few state-of-the-art LLMs and \oone{} reasoning models over different tasks in ACPBench. While the largest LLMs achieve more than $80\%$ accuracy on few tasks, the variance in performance across tasks and across LLMs is still big. This signifies the long way to go before they can be reliably used in practical scenarios.}
    \label{fig:intro}
\end{figure}

This work aims at evaluating and improving language models' ability to plan. However, end-to-end evaluation of planning ability is challenging. One, if an agent reaches a goal it does not necessarily mean it can plan. Second, evaluating a plan might be difficult in a domain where there can be multiple plans to achieve the goal.
So, instead of focusing on the entire end-to-end planning ability, we distill $7$ \atomic reasoning tasks 
that are critical for reliable planning
and
create datasets of such tasks. These tasks focus on reasoning about Actions, Change (transitions) and Planning; hence, we call our benchmark as ACPBench. The tasks include single step reasoning, like evaluating whether an action can be performed in the described state, as well as multi step reasoning, like whether a sequence of actions is a valid plan for the described state and the described goal.

For each task, ACPBench features both boolean (Bool) and
multiple-choice (MCQ) style questions from $13$ domains. All the datasets are generated from a formal representation of the domain in Planning Domain Definition Language (PDDL)~\cite{mcdermott-aimag2000}. Twelve of these domains are well-established benchmarks in both planning and reinforcement learning communities, readily available in PDDL format. Inspired by the shuffle task in BigBenchHard Suite~\cite{SuzgunSSGTCCLCZ23_bbh}, we have created an additional domain from scratch. The benefit of constructing the dataset from PDDL descriptions is twofold. First, it allows us to use existing planning tools and second, and arguably more important, it allows obtaining {\em provably correct} information for all the tasks. Natural language templates for these domains were carefully crafted by $5$ researchers. These templates and planning tools enable us to generate massive data for each task.

We evaluate performance of \oone{} reasoning model and \numofmodels{} state-of-the-art language models (including open-sourced  \texttt{Phi-3 128K}~\cite{Phi3}, \texttt{Mixtral 8x22B}~\cite{mixtral8x22b}, \texttt{LLAMA-3 70B}~\cite{dubey2024llama3herdmodels}, and a closed source \texttt{GPT-4o}~\cite{gpt4o}) on the ACPBench. We found that,  with Chain-of-Thought prompting (COT) ~\cite{wei-et-al-neurips2022} and 2-shot examples, \texttt{GPT-4o} was only able to achieve $78.40\%$ accuracy on MCQ questions in the ACPBench; with lowest accuracy of $52.50\%$ for 
the most difficult (validation) task. Similarly, \oone{} preview achieves accuracy of $87.31\%$ on average for the MCQ questions, with lowest accuracy of $63.08\%$ for the most difficult task. Figure~\ref{fig:intro} shows the overall performance of few selected models on all $7$ tasks of ACPBench.
To understand whether the smaller language models can improve their performance on these tasks, we fine-tune a language model on these tasks. The fine-tuning resulted in substantial improvements in performance across tasks and even demonstrated the ability to generalize to previously unseen domains.

In summary, our contributions are as follows:
\begin{itemize}
    \item We identify a collection of \numoftasks{} reasoning tasks required for efficient planning and introduce the first of its kind large-scale benchmark---ACPBench.
    \item We evaluate \oone{} reasoning models and \numofmodels{} state-of-the-art language models of different sizes on ACPBench.
    \item We finetune a 8B parameter model and show that the finetuned model performs on par with the large models.
    \item We conduct three ablations. First to understand effects of in-context example and COT. Second to investigate if tasks in ACPBench capture the plan generation ability. Third to understand how LLMs' abilities have progressed over time for ACPBench tasks.
\end{itemize}

%% file: background_relatedwork.tex
\section{Related Work and Background}

Recognizing the importance of evaluating reasoning and planning ability of LLMs,
various benchmarks have been proposed~\cite{agentBench,AgentBoard}. Most relevant to our work are the benchmarks that are generated from PDDL tasks.
\citet{he-etal-2023-exploring} proposed a natural language based question answering style dataset to evaluate LLMs on 4 tasks of projection, execution, planning, and goal recognition. 
PlanBench~\cite{valmeekam-et-al-neurips2023} is a benchmark suite with $8$ planning tasks including plan generation, reasoning about plan execution, and plan verification. 
Both these benchmarks focus on a limited number of planning domains (mainly the BlocksWorld domain), employing a template-based approach to generate natural language text.
In contrast, AutoPlanBench \cite{stein-et-al-arxiv2024} proposes to leverage LLMs to generate the natural language template. Specifically, they prompt an LLM for natural language template per predicate and per action. By reducing the human effort required for template generation, they were able to scale up the dataset to $12$ domains. However, they limit their focus to a single task - plan generation. 

\input{example}

In parallel, \citet{HandaActionReasoningBench2024} proposed ActionReasoningBench, featuring six tasks: 
 Fluent Tracking, State Tracking, Action Executability, Effects of Actions, Numerical RAC, and Composite Questions. Although there is some overlap between the tasks in ActionReasoningBench and ACPBench (for example, the Effects
 of Actions task overlaps with our Progression task), the majority of the tasks we propose are not covered by ActionReasoningBench:  Reachability, Action Reachability, Validation, Justification, Landmarks. Similarly, the following ActionReasoningBench tasks are not covered in ACPBench: State Tracking, and Numerical RAC. 

We now switch to providing the necessary  background. 
The ACPBench questions collection is generated based on PDDL tasks.
A PDDL task is defined over the first-order language; consisting of predicates, variables, and objects. A state $s$ is defined as a conjunction of grounded (by objects) predicates, also called atoms. An action $a$ is defined as a triple $\langle \pre(a), \add(a), \del(a) \rangle$; consisting of preconditions, add effects and delete effects, each being a conjunction of atoms. An action $a$ is applicable in a state $s$ if the state satisfies the preconditions of the action, i.e $pre(a) \subseteq s$. On performing an action $a$ in state $s$, the world transitions to the next state $t = s[a] = s \setminus \del(a) \cup \add(a)$. 
A goal $g$ is also a conjunction of atoms, and a state $s$ is a goal state if $g \subseteq s$. A sequence of actions $\pi_s = a_1\ldots a_n$ is applicable in the state $s$ if the actions are applicable in a sequence to the resulting states. $\pi_s$ is a plan for the state $s$ if $\pi_s$ is an applicable sequence of actions that results in a goal state.

%% file: example.tex
\begin{figure}[!t]
	\begin{minipage}{\columnwidth} 
 \begin{scriptsize}
\begin{lstlisting}[
linewidth=\columnwidth,
backgroundcolor =\color{lightgray!25},
    showspaces=false,                tabsize=1,showstringspaces=false,breakautoindent=false,breakindent=1ex]
Context: This is a swap domain where agents are swapping items or roles. Each agent is always assigned a single item/role. The goal is to obtain desired items/roles assigned. There are 8 agents: carol, michelle, xena, vic, dave, zoe, heidi, and alice. There are 8 items/roles: quadcopter, frisbee, necklace, whale, iceskates, guitar, zebra, and slinky. Currently, heidi is assigned necklace, michelle is assigned quadcopter, dave is assigned iceskates, vic is assigned whale, xena is assigned slinky, carol is assigned frisbee, alice is assigned zebra, and zoe is assigned guitar.
Bool: Is the following action applicable in this state: trade guitar of zoe for iceskates of dave?
MCQ: Which of the following actions will be applicable in this state? 
A. exchange frisbee of carol with zebra of alice. 
B. exchange guitar of zoe with necklace of vic. 
C. exchange guitar of heidi with zebra of zoe. 
D. exchange guitar of vic with zebra of zoe.
	\end{lstlisting}
 \end{scriptsize}
	\end{minipage} 
	\caption{ Example of boolean and multi-choice questions from the Applicablity task in ACPBench. The context contains the domain and the problem description. Query to LLM consists of context and a boolean or multi-choice question.}
	\label{fig:swapexample}
	\end{figure}

%% file: ACPBench.tex
\section{ACPBench}

\subsection{Domains}

ACPBench collection consists of $11$ classical planning domains, Alfworld~\cite{ALFWorld20}, and a novel swap domain. 
The $11$ classical planning domain, which were also used by AutoPlanBench \cite{stein-et-al-arxiv2024}, have public problem instance generators~\cite{seipp-et-al-zenodo2022}. 
Alfworld is a text-based reinforcement learning environment where an agent is given house hold tasks like `put a pan on the table' as a goal. Alfworld uses goals from the Alfred dataset~\cite{Alfred_ShridharTGBHMZF20} and encodes the dynamics of the domain as PDDL. This PDDL domain is publicly available\footnote{\url{https://github.com/alfworld/alfworld/blob/master/alfworld/data/alfred.pddl}} and PDDL problem files are obtained from the MINT benchmark~\cite{MINT}. 
For the novel Swap domain, we created the PDDL domain and the problem instance generator. Figure~\ref{fig:swapexample} contains an example problem description in this domain. All the domains are summarized in
Table~\ref{tab:domain_summary}.

We meticulously curated a set of templates to transform the PDDL task into a natural language description.
Following AutoPlanBench, we explored using LLMs to  automatically generate the templates, however, we found the templates were not reliable and needed significant modification. So, instead, 5 researchers crafted the translations,
carefully selecting and refining the templates to ensure they accurately convey the desired information. 
Specifically, we have templates for {\em domain description}, {\em problem description} and {\em actions}, from which we can compose (partial) states -- current state or a goal. 
These three templates, together with the PDDL files, are to be provided for every new domain, should we decide to extend the benchmark in the future.

\subsection{ACPBench Tasks}
We focus on $\numoftasks{}$ reasoning tasks within the realm of planning.
For each task, we provide a description and explain how the data was collected.

\begin{table}[t]

    \centering
    \begin{tabular}{c|r|r|r}
    Domain & \# Pred. & \# Actions & Max char.\\
    \midrule
Blocksworld & 5 & 4 & 1770\\
Logistics & 9 & 6 & 1065\\
Grippers & 4 & 3 & 1057 \\
Grid & 12 & 5 & 1235 \\
Ferry & 7 & 3 & 2132\\
FloorTile & 10 & 7 & 3196 \\
Rovers & 25 & 9 & 3631 \\
VisitAll & 3 & 1 & 1347 \\
\midrule
Depot & 6 & 5 & 1301\\
Goldminer & 12 & 7 & 1140 \\
Satellite & 8 & 5 & 4302\\
Swap & 1 & 1 & 849 \\
Alfworld & 34 & 19 & 4099\\
    \end{tabular}
    \caption{\footnotesize Statistics of the \numofdomains{} domains in ACPBench. The top \numoftraining{} domains are used for finetuning as well as evaluation. The bottom  \numoftesting{} domains are exclusively used for evaluations. Second column indicates the number of predicates in the PDDL domain, third column presents the number of lifted actions in the domain, and the last column indicates the max character length of the NL problem description in the generated dataset.}
    \label{tab:domain_summary}
\end{table}

\input{task_app}

\input{task_prog}

\input{task_reach}
\input{task_areach}
\input{task_val}

\input{task_just}

\input{task_land}

\subsection{Data Generation}

We use $25$ PDDL problem files of varying sizes per domain.
The specific arguments used to generate these problem files can be found in the appendix. These 25 tasks are partitioned into a training and a test set.
For each task, we use classical planners to generate a large collection of $1000$ plans~\cite{katz-lee-ijcai2023,katz-sohrabi-aaai2020}. With these plans, we sample the state space as follows. First, given a set of plans, we gather the states along these plans. Then, in order to obtain a diverse sample, we run random rollouts from each of the states found. The number of plans and the sample size are configurable parameters. In the {\em landmarks} task described above, we also find plans for the sampled states. To do that, we replace the initial state with the sampled state in the planning problem instance and run a top-k planner \cite{katz-lee-ijcai2023}. For finding mutexes, we exploit lifted mutex groups implementations from~\citet{fiser-aaai2020}. In this manner, we can potentially generate as many examples as we want. But to keep the test set of reasonable size, we generate only $10$ examples per domain, per task.\footnote{This test set will be made publicly available upon acceptance.}

%% file: task_app.tex
\paragraph{1. Applicability (\appl)}

The first, basic requirement for efficient planning is to determine the valid, available actions in a given situation. Various existing work have discussed LLMs fall short of this basic ability. When using GPT-4 Turbo for travel planning,~\citet{xie2024travelplanner} found that more than $30\%$ of the failed plans had \emph{invalid action dead loop}--that is even when the model was informed that the action is invalid, LLMs repeated these actions. 

For an action to be valid, its preconditions must hold in the state. 
Given a state $s$ and the set of actions $\actions$, the subset of applicable actions would be $\actions(s) = \{ a\in\actions \mid \pre(a)\subseteq s\}$, easily computable by iterating over the actions.
We therefore can create a boolean question with a positive answer by sampling from  $\actions(s)$ and with a negative answer by sampling from $\actions\setminus\actions(s)$. 
A multiple-choice question (MCQ) can be created by sampling the correct answer from $\actions(s)$ and wrong candidates from $\actions\setminus\actions(s)$. 
Figure~\ref{fig:swapexample} shows example of the domain description and problem description used in the context as well as one example each of Bool and MCQ question for applicability task.

%% file: task_prog.tex
\paragraph{2. Progression (\prog)} 

The next task evaluates LLMs ability to understand the outcome of an action or change. This ability is important to track information across transitions. 
The subpar performance of LLMs on the \emph{Tracking Shuffled Objects} task in the Big Bench Hard dataset suggests a significant limitation in their ability to reason about the consequences of actions or changes~\cite{SuzgunSSGTCCLCZ23_bbh}. 
Further, a few papers have proposed to use LLMs to execute a plan. For example, \citet{WangXLHLLL23_plannsolve} asks LLM to devise a plan and execute it step-by-step to reach the goal. To faithfully execute a plan, it is important for LLMs to demonstrate understanding of progression; how the world state is changed by the action. 

When a valid action is performed, the state changes in the following manner:
The delete effects of that action will no longer hold and the add effects will hold. Everything else remains unchanged. Given a state $s$ and an action $a$, the next state is $t=s\setminus \del(a)\cup \add(a)$. We can now partition the facts in the problem into four sets: the facts that held before applying the action and still hold ($s\cap t$), the facts that held before but not anymore ($s\setminus t$), those that did not hold but now hold ($t\setminus s$), and those that did not hold before and still don't hold ($F\setminus (s\cup t)$). While the answer of whether the fact is true after applying the action depends only on whether it is in $t$, the chain of thoughts leading to the answer differs for the aforementioned four cases.
We construct a boolean question by sampling from each of the four fact sets (if they are not empty), getting at most two positive and two negative examples per state. A single MCQ is constructed by sampling one possible answer from each of the four fact sets (non-empty ones), according to a uniform procedure described above.

%% file: task_reach.tex
\paragraph{3. Reachability (\reach)} 

The reachability task evaluates if a specific sub-goal can eventually be reached from the given state by taking (possibly multiple) actions. This is a multi-step reasoning task that can help avoid exploring unfeasible options.
To maximize the efficiency of LLMs, it is crucial to detect unreachable (sub)goals early on. This can avoid unnecessary prompting and wasteful exploration, ensuring that the LLMs are utilized effectively, especially when used during search \cite{yao-et-al-neurips2023}.

Reachability is PSPACE-hard to answer positively in general \cite{bylander-aij1994} for a specific fact, since that would require an evidence - a sequence of actions that achieves a state where the specified facts hold. However, generating positive examples is easy, based on any action sequence, taking the facts out of the end state. For negative examples, we explore multiple cases of unreachable facts and fact pairs. 
First, existing planning methods (under)approximate the reachability with poly-time computable delete-relaxed reachability \cite{hoffmann-nebel-ecp2001}. Facts that are not delete-relaxed reachable are therefore guaranteed not to be reachable. 
Another possible reason for a pair of facts that are individually reachable not to be reachable in the same state is if they are mutually exclusive \cite{lin-kr2004,fiser-komenda-jair2018}. A simple example of mutually exclusive facts in the ferry domain are {\em (empty-ferry)} and {\em (on ?c)}, meaning that the ferry cannot be empty and at the same time a car is on the ferry. 
Third, static facts that are not true in the initial state will never become true. For instance, c0 can never become a location, so {\em (location c0)} is unreachable (not captured by the methods in the first case, as they focus solely on non-static predicates). 
The chain of thoughts for a positive example is based on a sequence of actions that achieve the fact. For the negative examples, the chain of thoughts follows the argument laid out above for each of the cases. As in the previous case, the MCQ is captured by choosing from the lists of positive and negative options.

%% file: task_areach.tex
\paragraph{4. Action Reachability (\areach)} 

In API-driven workflows, the objective is typically presented as an instruction to execute a specific function~\cite{toolllm}. In these scenarios, an LLM must identify the necessary prerequisites for execution and formulate a strategy to meet them. Therefore, it is essential for LLMs to assess whether a given instruction is executable from the provided starting point. We formulate this ability as action reachability task.

The action reachability task is closely related to the atom reachability.
If an action model is available, then action reachability is equivalent to the atom reachability over the preconditions of the action. Therefore, this task requires an additional reasoning step about action preconditions. Similarly to the atom reachability task, the positive examples are generated from action rollouts, while the negative examples are generated by collecting actions with preconditions including unreachable atoms according to two of the three cases mentioned above delete-relaxed reachability and mutexes. The third case, unreachable static facts, was not used as often creates non-sensible actions {\em board car l0  at location c1}. Instead, we added incorrect action templates for each action, like {\em ``board the car c1 at location l0 into the airplane"} or {\em ``drive from location l0 to location l1"}.
Here as well, the chain of thoughts are created in a similar manner, and the MCQ is captured based on the positive and negative options lists.

%% file: task_val.tex
\paragraph{5. Validation (\validation)} 

A body of research has advocated the use of LLMs for validation and refinement~\cite{Reflexion_ShinnCGNY23,GouSGSYDC24_critic,MadaanTGHGW0DPY23_selfrefine}. In line with this research, we propose a Validation task. Here, given an initial state and a goal condition, the objective is to assess whether the specified sequence of actions is valid, applicable, and successfully achieves the intended goal.

There are essentially only four options in this case: (a) the sequence is not valid, (b) the sequence is valid, but not applicable, (c) the sequence is valid, applicable, but does not achieve the goal, and (d) the sequence is a plan. These are the four options used for all MCQ for this task. Since the options do not change, we generate four questions per sample, for each of the options to be a correct answer. In the boolean case, we create six different questions, with positive and negative variants for the three cases of whether the sequence is valid, applicable, and a plan. We generate the data for these questions from plans as follows. 
For the case (c), starting from a plan, we replace a suffix with a random rollout, ensuring that the goal is not achieved at the end of the rollout, but the sequence remains applicable. For the case (b), we try to replace an action on the sequence with an inapplicable action (one whose precondition does not hold in the state), starting from the end of the sequence. Once successful, we return the sequence ending with the inapplicable action. For the case of (a), we simply randomly choose an action on the sequence to replace its template with an incorrect action template, as in the previous task.

%% file: task_just.tex
\paragraph{6. Justification (\just)} 

A major criteria for plans to be considered reasonable is whether they include unnecessary actions. 
In the realm of LLMs and API workflows, it is desirable to avoid calling unnecessary APIs as well as reduce wasteful explorations. Hence, it would be of immense value if LLMs are able to identify whether an action is necessary. This corresponds to the justification task in planning literature.

The justification task reasons whether every action is actually needed on the plan. The problem was studied in the literature \cite{fink-yang-cscsi1992,salerno-et-al-kr2023} and found to be NP-hard in general. However, optimal plans are known to have all their actions being justified and checking whether a single action or a pair of consequent actions can be removed can be done in polynomial time.
We consider the following cases, for either a single action or a pair of consequent actions in a plan: 1) a single action can be removed from the plan and the remaining plan is still a valid plan for the same problem 2) an action cannot be removed from the plan 3) the consequent pairs of actions can be removed from the plan 4) the immediate pairs of action cannot be removed from the plan. Note that we truncate the considered plans and only consider two actions after the goal is reached except if the truncation leads to a non-plan. 
Given a large set of plans, we consider the above four cases, and generate positive and negative examples for both boolean and multiple choice questions.

%% file: task_land.tex
\paragraph{7. Landmarks (\landmark)} 

LLMs have shown to hallucinate or deviate from the task when the trajectory is long~\cite{Huang_survey}. To alleviate this problem, various work has proposed to use LLMs to decompose the goal into subgoals and achieve each of these subgoals separately.
To do this faithfully, it is crucial for LLMs to be able to identify subgoals that are necessary to achieve the goal. In planning literature such subgoals are often called landmarks \cite{porteous-et-al-ecp2001}. Landmarks are facts that must become true sometime along every plan. So, the last task in ACPBench evaluates LLMs ability to recognize landmarks.

While checking whether a fact is a landmark is PSPACE-hard \cite{porteous-et-al-ecp2001}, there are several methods that can find a subset of landmarks~\cite{keyder-et-al-ecai2010,hoffmann-et-al-jair2004,richter-et-al-aaai2008,zhu-givan-icaps2003dc}. We use the so-called RHW method~\cite{richter-et-al-aaai2008}. Further, negative evidence can be obtained from a collection of plans - a fact that does not appear on all of these plans is not a landmark. We sample from positive and negative examples obtained that way and construct two boolean questions and one MCQ. Here as well, the chain of thoughts generated capture the described logic.

%% file: experiments.tex
\section{Experiments}
\label{exp}

\subsection{Evaluation of pre-trained language models}

\input{compare_models}

We first analyse how existing pre-trained language models perform on ACPBench. Table~\ref{tab:compare_models} presents the accuracy of all the language models on the \numoftasks{} ACPBench tasks. These results are mean over $5$ runs for all models; except GPT family models and LLAMA-3.1 405B~\cite{dubey2024llama3herdmodels}, which were run once due to resource constraints. All LLMs were either accessed using 
API or hosted locally using hugging face transformer library on machines with 2 $A100$ $80$GB GPU.  Note that accuracy of $50.00$ on boolean questions indicates that the performance of the model is as good as a random guess. As all the MCQs in the datasets have $4$ options, accuracy less than $25.00$ indicates that the performance is worse than random guess. 
To investigate the out-of-the-box performance, we restrict the evaluation to single turn Chain-of-Thought (COT) prompting with two in-context examples. An example prompt for the Bool applicability question
is shown in Figure~\ref{fig:example_prompt}. 

\input{example_prompt}

Notably, \texttt{LLAMA-3.1 405B} and \texttt{GPT-4o} consistently outperform other models on these tasks, although they do not \textbf{always} achieve the top performance. When it comes to smaller open-sourced models, \texttt{Codestral 22B} stands out for its exceptional performance on boolean questions, while \texttt{Mixtral 8x7B} excels in handling multi-choice questions. However, both of them lag significantly behind \texttt{GPT-4o}, which is the best performer in these tasks. Action Reachability and Validation are the most challenging tasks for LLMs. Surprisingly, the GPT family models are not even among top-3 for the action reachablity task. Across all the tasks, GPT-4o performs best for boolean questions and LLAMA-3.1 405B performs best for multi-choice questions.

\begin{figure}[!t]
    \centering
    \includegraphics[width=\columnwidth]{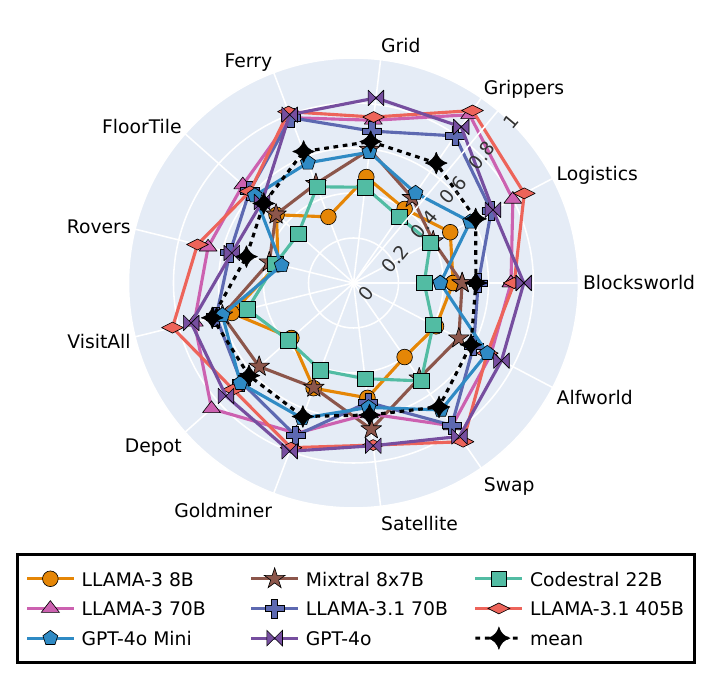}
    \caption{\footnotesize Comparison of $8$ top performing LLMs on multi-choice questions in
    \numofdomains{} domains of ACPBench. The mean of performance across the top-$8$ models is presented with dotted line in Black. The mean line indicates that none of the domains are exceptionally easy.}
    \label{fig:domain_analysis}
    \vspace{-2em}
\end{figure}

Figure~\ref{fig:domain_analysis} displays a domain-wise analysis of the performance of LLMs on multi-choice questions. This analysis showcases the top $8$ performing models\footnote{All supplementary information on the remaining models and boolean questions are relegated to the Appendix to maintain clarity.}. The average performance of these top-$8$ models is shown in Figure~\ref{fig:domain_analysis} as the dotted line in black. This indicates that 
across models no specific domain seems too easy. However, Rovers, FloorTile, Blocksworld, Alfworld and Satellite domains pose the greatest challenges to LLMs, in that particular order.

\subsection{Fine-tuning}
Foundational models, and LLMs specifically, have shown to improve performance on specific tasks when they are finetuned for those tasks. So, next we investigate if finetuning a language model provides any improvement. For this investigation, we keep aside the following 5 domains, Depot, Goldminer, Satellite, Swap, and Alfworld, and generate a training set for the remaining 8 domains. Then we pick one of the small models, \texttt{Granite-code 8B} \cite{granite_code}, and finetune it with QLoRA. The resulting performance improvement is shown in Table~\ref{tab:finetuned_train}. As finetuned models have already seen examples during training, we use only IO prompts with the finetuned model. We finetuned Granite 8B available on HuggingFace on a machine with two A100 80GB GPUs. 

Upon finetuning, the average accuracy of the model improves from $51.43\%$ to $95.71\%$ on boolean questions and from $19.18\%$ to $94.29\%$ on multi-choice questions. Further, Table~\ref{tab:finetuned_test} presents the performance on the remaining \numoftesting{} unseen domains. 
It is remarkable to observe such a significant improvement even on unseen domains; sometimes surpassing the GPT-4o performance. This indicates that finetuning a model, even on a separate domain, improves performance on these tasks. The right-most column in Tables~\ref{tab:finetuned_train} and \ref{tab:finetuned_test} presents the performance of the best on that task LLM with COT 2-shots prompting. As can be seen; \texttt{Granite Finetuned} model outperforms the best of all models for most of the tasks in the training domains. Even in testing domains, the accuracy difference is significantly reduced upon finetuning.

\subsection{Ablations}

\subsubsection{Prompt Style}
From previous section, it is clear that COT 2-shot yields better results than IO prompts for ACPBench tasks. However, it is not clear whether COT or 2-shot examples provide the performance gain. To investigate this, we perform the following ablation study. We compare four prompt styles: (1) IO prompt, (2) Chain-of-Thought prompt without in-context examples (COT), (3) IO prompt with two in-context examples (IO 2-shots), and (4) Chain-of-Thought with two in-context examples (COT 2-shots).\footnote{Examples of prompts are included in the Appendix.}

\input{finetuning}

\begin{figure}[!th]
    \vspace{-1em}
    \centering
    \includegraphics[width=\columnwidth]{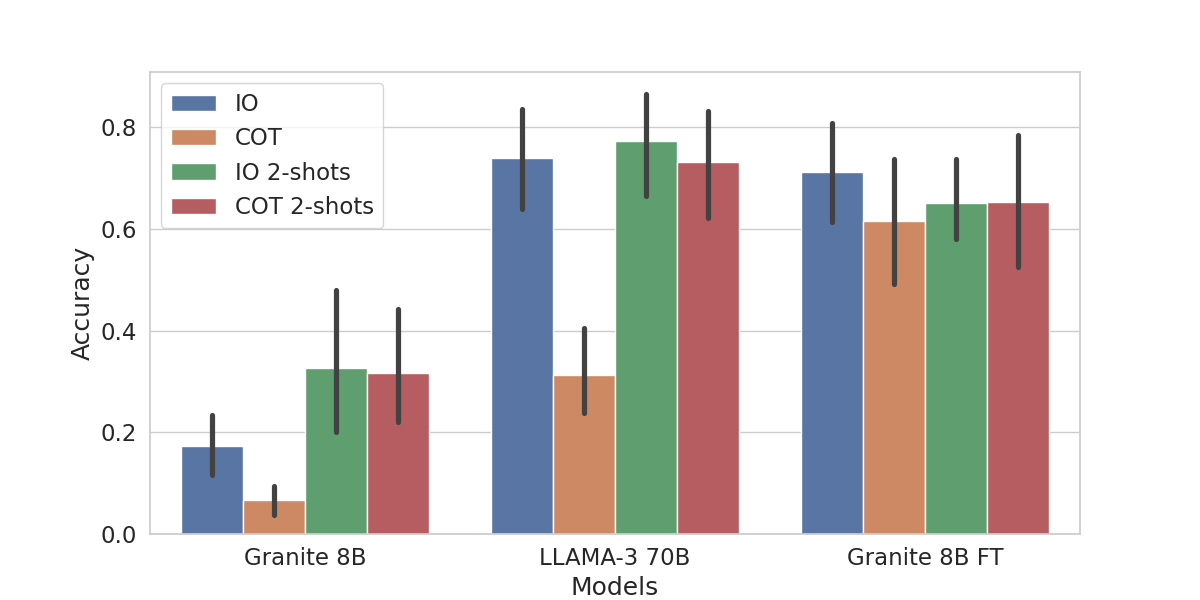}
    \caption{\footnotesize Comparison of different prompt styles on two pretrained models: Granite 8B and LLAMA-3 70B, and finetuned Granite 8B model for MCQ tasks in \numoftesting{} testing domains.}
    \label{fig:promptstyle}
\end{figure}

We include \texttt{Granite-code 8B base} model, \texttt{LLAMA-3 70B} (one of the top-performing open source model), and the \texttt{Granite-code 8B finetuned FT} model. To have a fair comparison, we use 2-shot examples from the training domains and only compare performance on the testing domains for MCQ tasks. Figure~\ref{fig:promptstyle} presents the results. For the two pretrained models, we see that while COT 2-shots prompting yields better result than IO, IO 2-shots prompting had the best performance. For finetuned model, we see that neither COT nor 2-shots provide any advantage; rather IO prompts yield the best results.

\subsubsection{Generalization}

ACPBench consists of tasks that are crucial for effective, robust and reliable planning. Improving performance on ACPBench should improve LLM's ability to reason about these tasks, and hence should improve LLM's ability to generate plans. To verify this hypothesis, we compare the \texttt{Granite-code Base 8B} model and \texttt{Granite-code finetuned 8B} model on plan generation task (t1) in PlanBench~\cite{valmeekam-et-al-neurips2023datasets}. Table~\ref{tab:planbench} presents the results. Granite finetuned model, which was QLoRA~\cite{DettmersPHZ23_QLoRA} trained on ACPBench tasks for \numoftraining{} training domains, shows improvement on plan generation ability.

\input{planbench}

\begin{figure}
    \centering
    \includegraphics[width=\columnwidth]{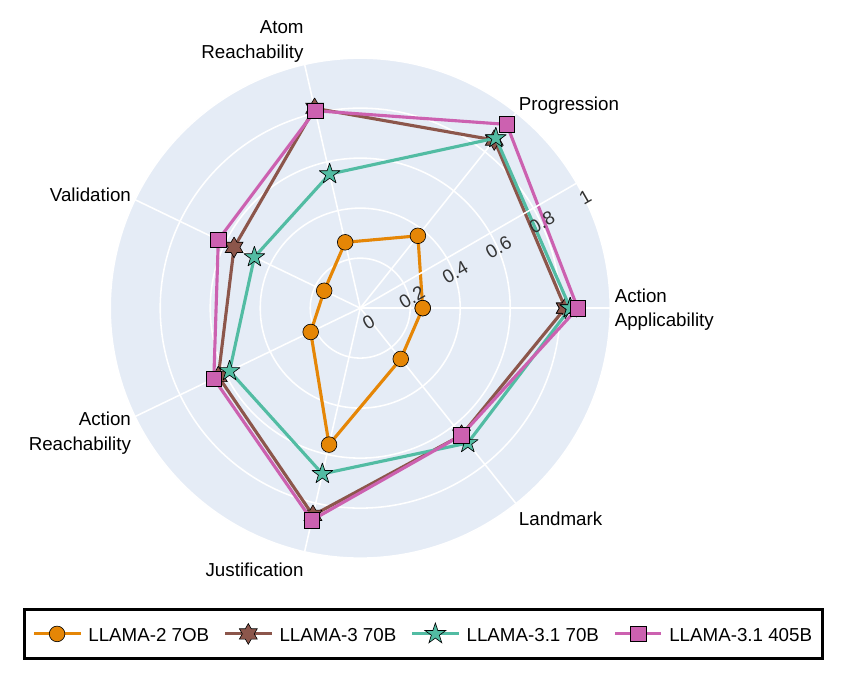}
    \caption{Comparison of how various versions of LLAMA perform on ACPBench tasks.}
    \label{fig:llama}
    \vspace{-2em}
\end{figure}

\subsubsection{Performance over time}

One of our major motivations to generate and release the ACPBench collection is to encourage researchers to address the poor performance on these tasks and build models that are capable to perform reasoning required for better planning. We believe that without such benchmarks, the progress toward this goal is ad-hoc. To verify our belief, we perform a small ablation, where we compare LLAMA family of models that were released over last $6$ months to see if there is any improvement on ACPBench tasks over time. Figure~\ref{fig:llama} presents the performance of \texttt{LLAMA-2 (70B), LLAMA-3 (70B)} and \texttt{LLAMA-3.1 (70B and 405B)} on ACPBench tasks. We see that there is a significant jump in performance between \texttt{LLAMA-2} and \texttt{LLAMA-3}. However, the difference in performance of \texttt{LLAMA-3 70B} and \texttt{LLAMA-3.1 405B} is not significant. This highlights the need for benchmarks that systematically captures the reasoning ability required for planning. 

%% file: compare_models.tex
\begin{table*}[t]
    \centering
    \resizebox{\textwidth}{!}{
    \begin{tabular}{l@{}|r@{\hspace{0.6\tabcolsep}}r|r@{\hspace{0.6\tabcolsep}}r|r@{\hspace{0.6\tabcolsep}}r|r@{\hspace{0.6\tabcolsep}}r|r@{\hspace{0.6\tabcolsep}}r|r@{\hspace{0.6\tabcolsep}}r|r@{\hspace{0.6\tabcolsep}}r|r@{\hspace{0.6\tabcolsep}}r}
    \multirow{2}{*}{\textbf{Model}}  & \multicolumn{2}{c|}{Applicability} & \multicolumn{2}{c|}{{Progression}}&  \multicolumn{2}{c|}{Reachability} & \multicolumn{2}{c|}{Validation} & \multicolumn{2}{c|}{Action Reach.} &  \multicolumn{2}{c|}{Justification} & \multicolumn{2}{c|}{Landmark} & \multicolumn{2}{c}{Mean}\\
 & Bool & MCQ & Bool & MCQ& Bool & MCQ& Bool & MCQ& Bool & MCQ& Bool & MCQ& Bool & MCQ& Bool & MCQ\\
 \midrule
    Phi-3 128K& $66.15$& $33.08$& $68.46$& $53.85$& $52.31$& $26.15$& $50.77$ & $19.23$& $53.33$& $32.50$& $49.23$& $33.85$& $49.23$& $46.92$ & $55.53$ & $34.75$ \\
Gemma 7B & $63.23$& $28.62$& $64.92$& $31.08$& $53.08$& $23.08$& $46.92$ & $20.0$& $55.67$& $34.50$& $50.77$& $36.46$& $27.54$& $30.31$ &$51.80$ & $28.93$ \\
Granite 7B & $56.92$& $29.54$& $55.23$& $35.38$& $50.77$& $34.62$& $32.15$ & $26.15$ & $48.33$& $28.33$& $40.77$& $25.38$& $47.69$& $32.15$ & $48.20$ & $29.67$ \\
Mistral 7B & $61.54$& $32.31$& $73.08$& $38.46$& $53.08$& $28.46$& $47.85$ & $17.69$& $\best{\bestofmini{65.00}}$ & $19.17$& $48.46$& $30.00$& $35.38$& $33.08$ & $55.00$ & $28.67$ \\
Mistral Inst. 7B & $63.08$& $31.54$& $61.54$& $46.92$& ${61.54}$& $33.08$& $52.15$ & $36.15$& $45.83$& $34.17$& $43.08$& $29.23$& $57.69$& $50.77$ & $55.45$ & $37.30$ \\
Granite-c 8B  & $59.23$& $32.31$& $70.00$& $34.31$& $52.31$& $24.31$& $44.15$ & $17.08$& $57.50$& $25.83$& $46.92$& $34.62$& $37.23$& $35.38$ & $53.09$ & $29.21$ \\
Granite-c Inst. 8B& $55.38$& $32.31$& $69.23$& $34.46$& $50.77$& $29.23$& $45.85$ & $22.31$ & $42.50$& $39.33$& $46.15$& $32.31$& $43.85$& $38.46$ & $50.53$ & $32.63$ \\
LLAMA-3 8B & $72.92$& $49.23$& $73.08$& $56.00$& $55.23$& $41.08$& $51.54$ &  $\bestofmini{{49.23}}$& $\secondbest{63.50}$& $36.67$& $57.54$& $32.31$& $56.92$& $43.85$ & $61.53$ & $44.05$ \\
LLAMA-3.1 8B & $65.38$& $56.92$& $63.85$& $47.69$& $53.08$& $33.85$& $60.00$ & $37.69$& $42.50$& $28.33$& $46.92$& $45.38$& $33.85$& $40.00$ & $51.46$ & $41.52$ \\
Mixtral 8x7B & $75.85$& $\bestofmini{57.69}$& $74.00$& $\bestofmini{61.38}$& $\bestofmini{76.00}$& $40.00$& $65.69$ & $34.77$& $52.83$& ${\bestofmini{55.00}}$& $55.38$& $51.38$& $59.54$& $\bestofmini{60.00}$ & $65.53$ & $\bestofmini{51.44}$ \\
Granite 13B & $42.00$& $29.23$& $52.46$& $20.77$& $47.69$& $28.46$& $51.54$ & $34.62$ & $45.17$& $26.33$& $45.38$& $27.69$& $50.31$& $19.23$ & $47.79$ & $26.66$ \\
Codestral 22B & $\bestofmini{84.62}$& $39.23$& $\bestofmini{83.85}$& $51.54$& $54.62$& $28.46$& ${\bestofmini{66.15}}$& $24.62$& $53.33$& $38.33$& $\bestofmini{67.69}$& $\bestofmini{62.31}$& $59.23$& $42.31$ & $\bestofmini{67.4}$ & $40.97$ \\
Mixtral 8x22B & $80.77$& $37.69$& $72.31$& $54.62$& $50.00$& $\bestofmini{42.62}$& $37.69$ & $16.92$& ${58.50}$& $27.83$& $43.08$& $44.62$& $44.77$& $45.23$ & $55.63$ & $39.25$ \\
Deepseek Inst. 33B & $70.77$& $37.23$& $68.46$& $46.31$& $53.08$& $31.69$& $51.54$ & $37.69$& $50.00$& $27.50$& $46.92$& $26.15$& $\bestofmini{62.31}$& $39.23$ & $57.58$ & $35.11$ \\
LLAMA-c 34B & $80.77$& $42.31$& $73.08$& $43.85$& $53.08$& $25.69$&$50.15$ & $28.46$ & $53.17$& $33.33$& $55.38$& $35.38$& $46.92$& $40.62$ & $59.02$ & $35.71$ \\

\midrule
LLAMA-2 7OB & $78.46$& $24.62$& $71.54$& $36.77$& $53.08$& $26.92$& $51.38$ & $16.15$& ${60.83}$& $22.00$& $49.23$& $55.54$& $24.46$& $26.00$ & $55.72$ & $29.71$ \\
LLAMA-c 70B & $74.77$& $36.15$& $54.77$& $52.92$& $48.62$& $23.69$& $40.0$ & $17.69$& $49.67$& $28.83$& $46.92$& $31.54$& $37.08$& $42.31$ &  $50.9$ & $32.87$ \\
LLAMA-3 70B & ${90.77}$& ${82.31}$& $93.08$& ${86.15}$& $\best{87.69}$& $\best{82.31}$& 
$\best{78.62}$& $\secondbest{56.62}$& ${60.50}$& $\secondbest{63.00}$& $62.31$& $\secondbest{85.38}$& ${78.15}$& $64.77$ & ${78.71}$ & ${74.30}$ \\
LLAMA-3.1 70B & $93.08$ & $84.31$ & $89.85$ & $86.77$ & $61.38$ & $54.92$ & $66.15$ & $46.62$ & $63.00$ & $58.00$ & $56.92$ & $68.46$ & $34.62$ & $\secondbest{69.23}$ & $66.67$ & $66.94$ \\
LLAMA-3.1 405B & $\secondbest{95.38}$ & $\secondbest{86.92}$ & $93.08$ & $\best{93.85}$ & $59.23$ & $\secondbest{80.77}$ & $\secondbest{77.23}$ & $\best{62.92}$ & $\best{65.00}$ & $\best{65.00}$ & $\best{90.00}$ & $\best{86.92}$ & $\secondbest{83.08}$ & $65.38$ & $\secondbest{80.49}$ & $\best{77.42}$ \\
\midrule
\gptmini & ${90.77}$& $73.85$& $\best{95.38}$& $79.23$& $\secondbest{80.77}$& $39.23$& $67.69$ & $46.15$ & $54.17$& $21.67$& ${77.69}$& $70.00$& $76.92$& ${67.69}$ & $77.74$ & $56.50$ \\

\gpt & $\best{96.92}$ & $\best{89.23}$& $\secondbest{94.62}$& $\secondbest{90.00}$& $79.23$& ${76.92}$& $61.54$ & $53.85$& $57.50$& ${52.50}$& $\secondbest{88.46}$& ${80.77}$& $\best{95.38}$& $\best{79.23}$  &$\best{81.84}$ & $\secondbest{74.97}$ 
    \end{tabular}
    }
    \caption{Accuracy of \numofmodels{} LLMs on \numoftasks{} ACPBench tasks (boolean as well as multi-choice questions). The best results are $\mathbf{boldfaced}$, second best are $\secondbest{underlined}$, and the best among the small, open-sourced models are highlighted with $*$. All models were evaluated with two in-context examples and Chain-of-Thought prompt. The right-most column is mean across tasks.}
    \label{tab:compare_models}
\end{table*}

%% file: example_prompt.tex
\begin{figure}[!t]
	\begin{minipage}{\columnwidth} 
  \begin{scriptsize}
\begin{lstlisting}[
backgroundcolor =\color{lightgray!25},
    showspaces=false,                
    showstringspaces=false,               tabsize=1,showstringspaces=false,breakautoindent=false,breakindent=1ex]
**Question**: This is a ferry domain, where the task is to transport cars from their start to their goal locations, using a ferry. Each location is accessible by ferry from each other location. The cars can be debarked or boarded, and the ferry can carry only one car at a time.  There are 2 locations and 2 cars, numbered consecutively.  Currently, the ferry is at l0, with the car c1 on board. The cars are at locations as follows: c0 is at l0.  Is the following action applicable in this state:  travel by sea from location l1 to location l0?   
**Thoughts**: Let's think step by step.  
Step 1: In order to apply the action travel by sea from location l1 to location l0, the following fact(s) must hold in this state: The ferry is at l1 location.
Step 2: These facts do not hold in the mentioned state. 
So, the action is not applicable. 
**Final Answer**: No.  
**Question**: ...
**Thoughts**: ...
**Final Answer**: Yes.
**Question**:  <context> + <question>
**Thoughts**: Let's think step by step.
    	\end{lstlisting}
      \end{scriptsize}
	\end{minipage} 
	\caption{Example of the COT prompt.}
	\label{fig:example_prompt}
	\end{figure}

%% file: finetuning.tex
\begin{table}[!t]
    \centering
    \resizebox{\columnwidth}{!}{
    \begin{tabular}{c@{}c|r|c@{}c|c@{}c|r}
    \multicolumn{2}{c|}{\textbf{Task}} & \textbf{Base IO} & \multicolumn{2}{c|}{\bf Base COT 2-shot} &\multicolumn{2}{c|}{\bf Finetuned IO} & \textbf{Best}\\
    \midrule
\multirow{2}{*}{\appl}& B & $53.75$ & $62.5$ & $(+8.75)$ & $\textbf{98.75}$ & $(+45.0)$ & $97.50$ \\  
      & MC & $15.0$       & $36.75$ & $(+21.75)$ & $\textbf{92.5}$ &  $(+77.5)$  & $90.00$\\  
\midrule
\multirow{2}{*}{\prog}& B & $52.5$ & $76.25$ & $(+23.75)$ & $\textbf{97.5}$ & $(+45.0)$ & $96.25$  \\  
     & MC & $22.5$       & $33.25$ & $(+10.75)$ & $\textbf{93.75}$ &  $(+71.25)$ & $\textbf{93.75}$ \\  
\midrule
\multirow{2}{*}{\reach}& B & $47.5$ & $52.5$ & $(+5.0)$ & $\textbf{97.5}$ & $(+50.0)$  & $87.50$ \\  
     & MC & $15.0$       & $20.75$ & $(+5.75)$ & $\textbf{98.75}$ &  $(+83.75)$  & $82.5$  \\  
\midrule
\multirow{2}{*}{\validation}& B & $45.0$ & $40.5$ & $(-4.5)$ & \textbf{$100.0$} & $(+55.0)$ & $78.75$ \\  
     & MC &  $38.5$       & $20.0$ & $(-18.5)$ & $\textbf{87.5}$ &  $(+49.0)$  & $57.75$ \\  
\midrule
\multirow{2}{*}{\areach}& B & $45.0$ & $56.25$ & $(+11.25)$ & $\textbf{97.5}$ & $(+52.5)$ & $65.75$ \\  
     & MC & $14.25$       & $28.75$ & $(+14.5)$ & $\textbf{95.0}$ &  $(+80.75)$  & $78.75$ \\  
\midrule
\multirow{2}{*}{\just}& B & $56.25$ & $50.0$ & $(-6.25)$ & $\textbf{97.5}$ & $(+41.25)$  & $90.0$\\  
      & MC& $16.25$       & $35.0$ & $(+18.75)$ & $\textbf{96.25}$ &  $(+80.0)$ & $82.5$ \\  
\midrule
\multirow{2}{*}{\landmark}& B & $60.0$ & $41.25$ & $(-18.75)$ & $81.25$ & $(+21.25)$ & $\textbf{97.50}$ \\  
      & MC& $20.0$       & $18.5$ & $(-1.5)$ & $\textbf{90.0}$ &  $(+70.0)$ & $71.25$ \\  
\midrule
\multirow{2}{*}{Mean} & B & $51.43$ & $54.18$ & $(+2.75)$ & $\textbf{95.71}$ & $(+44.28)$ & $81.07$\\
& MC & $20.21$ & $27.57$ & $(+7.36)$& $\textbf{93.39}$ & $(+73.18)$ & $77.68$ \\

    \end{tabular}
    }
\vspace{-0.2cm}
    \caption{\footnotesize Comparison of the \texttt{Granite-code Base 8B}  model and the finetuned model on \numoftraining{} training domains of ACPBench. We present accuracy values for the Base model with Input-Output prompts (IO) as well as with Chain-of-Thought prompt with two in-context examples (COT 2-shot). The values enclosed in parentheses represent the improvement over the base model w/ IO prompts. The right-most column presents the performance of the best LLM with COT 2-shot on training domain. Best results are in bold.}
    \label{tab:finetuned_train}
\end{table}

\begin{table}[!ht]
    \centering
    \resizebox{\columnwidth}{!}{
    \begin{tabular}{c@{}c|c|c@{}c|c@{}c|c}
    \multicolumn{2}{c|}{\textbf{Task}} & \textbf{Base IO} & \multicolumn{2}{c|}{\bf Base COT 2-shot} &\multicolumn{2}{c|}{\bf Finetuned IO}  & \textbf{Best} \\
    \midrule
\multirow{2}{*}{\appl}& B & $50.0$ & $54.0$ & $(+4.0)$ & $74.0$ & $(+24.0)$ & $\textbf{96.00}$ \\  
      & MC& $14.0$       & $25.2$ & $(+11.2)$ & $62.0$ &  $(+48.0)$ & $\textbf{88.00}$\\  
    \midrule
\multirow{2}{*}{\prog}& B & $50.0$ & $60.0$ & $(+10.0)$ & $80.0$ & $(+30.0)$ & $\textbf{94.00}$  \\  
      & MC& $28.0$       & $36.0$ & $(+8.0)$ & $82.0$ &  $(+54.0)$  & $\textbf{96.0}$\\  
    \midrule
\multirow{2}{*}{\reach}& B & $46.0$ & $52.0$ & $(+6.0)$ & $82.0$ & $(+36.0)$  & $\textbf{88.00}$ \\  
      & MC& $10.0$       & $30.0$ & $(+20.0)$ & $56.0$ &  $(+46.0)$  & $\textbf{82.00}$\\  
    \midrule
\multirow{2}{*}{\validation}& B & $46.0$ & $50.0$ & $(+4.0)$ & $80.0$ & $(+34.0)$  & $\textbf{84.0}$  \\  
     & MC & $26.0$       & $12.4$ & $(-13.6)$ & $54.0$ &  $(+28.0)$  & $\textbf{71.2}$ \\  
    \midrule
\multirow{2}{*}{\areach}& B & $35.0$ & $60.0$ & $(+25.0)$ & $\textbf{82.5}$ & $(+47.5)$  & $77.5$ \\  
      & MC& $5.0$       & $20.0$ & $(+15.0)$ & $\textbf{70.0}$ &  $(+65.0)$  & $57.50$ \\  
    \midrule
\multirow{2}{*}{\just}& B & $42.0$ & $42.0$ & $(+0.0)$ & $\textbf{98.0}$ & $(+56.0)$  & $96.0$ \\  
      & MC& $16.0$       & $34.0$ & $(+18.0)$ & $80.0$ &  $(+64.0)$ & $\textbf{94.0}$ \\  
    \midrule
\multirow{2}{*}{\landmark}& B & $44.0$ & $30.8$ & $(-13.2)$ & $72.0$ & $(+28.0)$ & $\textbf{92.0}$ \\  
      & MC& $20.0$       & $62.4$ & $(+42.4)$ & $92.0$ &  $(+72.0)$  & $\textbf{94.0}$ \\  
      \midrule
      \multirow{2}{*}{Mean}& B & 44.71 &  49.83 & (+5.21) & 81.21 & (+36.5) & $\textbf{82.79}$ \\
      & MC & 17.00 & 31.43 & (+14.43) & 70.86 & (+53.86) & $\textbf{78.07}$
    \end{tabular}
    }
    \vspace*{-0.2cm}

    \caption{\footnotesize Comparison of \texttt{Granite-code Base 8B} and finetuned model on \numoftesting{} ACPBench domains that are unseen during training. The values enclosed in parentheses represent the improvement over the base model w/ IO prompts. The right-most column presents performance of the best LLM with COT 2-shot on testing domain. Best results are in bold.}
    
    \label{tab:finetuned_test}
\end{table}

%% file: planbench.tex
\begin{table}[!t]
    \centering
    \begin{tabular}{c|c|c|c}
         Domain & Base & Finetuned & LLAMA-3 70B\\
         \midrule
         Blocksworld (600) & 24 & 44 & 57\\
         \midrule
         Logistics (285) & 14 & 15 & 14 
    \end{tabular}
    \caption{Comparison of \texttt{Granite-code Base}, \texttt{finetuned}, and \texttt{LLAMA-3 70B} model on PlanBench Dataset.}
    \label{tab:planbench}
\end{table}

%% file: o1_experiments.tex
\subsection{Reasoning Model: OpenAI o1}

\begin{figure*}
    \centering
    \begin{subfigure}[t]{0.5\textwidth}
        \centering
        \includegraphics[height=18em]{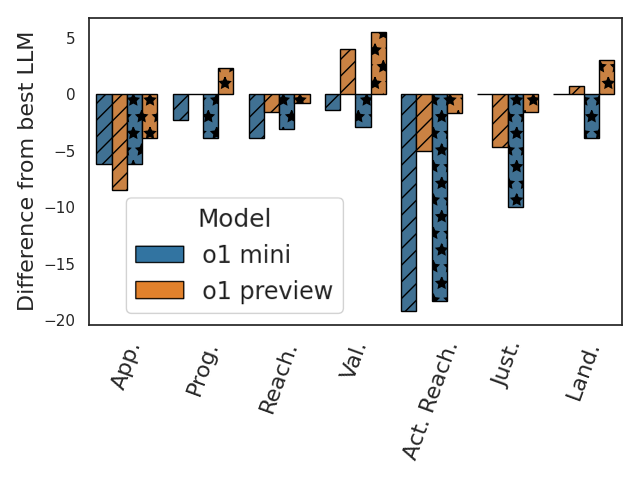}
        \caption{Bool}
    \end{subfigure}%
    \hfill
    \begin{subfigure}[t]{0.5\textwidth}
        \centering
        \includegraphics[height=18em]{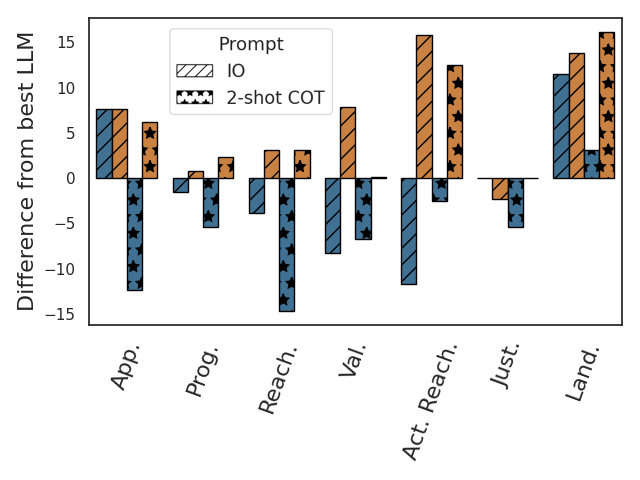}
        \caption{MCQ}
    \end{subfigure}
    \caption{Comparing \oone{} models with the best LLM. Positive difference shows \oone{} model performing better than the best of the LLMs. Negative difference is when \oone{} model lags behind the best LLM.}
            \label{fig:o1_diff}
\end{figure*}

\input{o1_compare}

Recently, OpenAI released a series of LLM-based reasoning models called \oone~\cite{o1}, that show significant improvement over GPT-4o on benchmarks that require reasoning. Although \oone{} preview and mini are made available via similar APIs as previous LLMs, 
they do not truly fit the LLM category; rather, they are a system (or an agent) that makes multiple calls to LLMs before providing an answer. Note that \oone were also referred to as Large Reasoning Models~\cite{valmeekam2024llmscantplanlrms}.
While we acknowledge the difference, it is interesting to compare the best performing LLMs to the \oone{} models. The comparison is depicted in 
Table~\ref{tab:compare_o1}. 
Further, Figure~\ref{fig:o1_diff} shows the performance difference of \oone{} models (with zeroshot IO and 2-shot COT prompts) from the best performing LLMs. Our results indicate that \oone{} models fail to yield performance gains for boolean questions, but demonstrate notable improvements on MCQs. Specifically, \oone{} preview consistently performs better or equal to the best performing model for MCQs. 
The responses for MCQ tasks suggests that \oone{} models consider each option individually, perform a case-by-case analysis, and only then select an option.

We would like to reiterate that while we present results of \oone{} side by side with LLAMA-3.1 405B and GPT-4o LLMs, the comparison is not even-handed due to below mentioned reasons:
\begin{itemize}
    \item All our LLM experiments had a generated token limit of $1024$; \oone{} models did not have that limit. On average the number of  tokens generated by \oone{} preview for MCQ tasks, where we see the maximum improvement, was $5705$ (this includes the completion token ($3164$) and the reasoning token ($2542$)).
    \item LLM evaluations are based on a single generation. We did not evaluate multi-turn prompts (such as self-consistency or self-reflection). \oone{} models seem to internally make multiple calls to an LLM.
\end{itemize}
In terms of pure monetary cost, the \oone{} evaluation is approximately $20$ times more expensive than of GPT-4o. It remains to be seen if a multi-turn prompting of an open-sourced LLM like LLAMA-3.1 can achieve similar improvement with lower cost.

%% file: o1_compare.tex
\begin{table*}[!ht]
    \centering
    \resizebox{\textwidth}{!}{
    \begin{tabular}{l@{}|r@{\hspace{0.6\tabcolsep}}r|r@{\hspace{0.6\tabcolsep}}r|r@{\hspace{0.6\tabcolsep}}r|r@{\hspace{0.6\tabcolsep}}r|r@{\hspace{0.6\tabcolsep}}r|r@{\hspace{0.6\tabcolsep}}r|r@{\hspace{0.6\tabcolsep}}r|r@{\hspace{0.6\tabcolsep}}r}
    \multirow{2}{*}{\textbf{Model}}  & \multicolumn{2}{c|}{Applicability} & \multicolumn{2}{c|}{{Progression}}&  \multicolumn{2}{c|}{Reachability} & \multicolumn{2}{c|}{Validation} & \multicolumn{2}{c|}{Action Reach.} &  \multicolumn{2}{c|}{Justification} & \multicolumn{2}{c|}{Landmark} & \multicolumn{2}{c}{Mean}\\
 & Bool & MCQ & Bool & MCQ& Bool & MCQ& Bool & MCQ& Bool & MCQ& Bool & MCQ& Bool & MCQ& Bool & MCQ\\
 \midrule
 \multicolumn{17}{c}{2-shot Chain-of-Thought prompt}\\
 \midrule
LLAMA 405B & {$95.38$} & $86.92$ & $93.08$ & $93.85$ & $59.23$ & $80.77$ & $77.23$ & $62.92$ & $65.00$ & $65.00$ & $90.00$ & $86.92$ & $83.08$ & $65.38$ & $80.43$ & $77.39$ \\
GPT-4o Mini & $90.77$ & $73.85$ & $95.38$ & $79.23$ & $80.77$ & $39.23$ & $67.69$ & $46.15$ & $54.17$ & $21.67$ & $77.69$ & $70.00$ & $76.92$ & $67.69$ & $77.63$ & $56.83$ \\
GPT-4o & {$96.92$} & $89.23$ & $94.62$ & $90.00$ & $79.23$ & $76.92$ & $61.54$ & $53.85$ & $57.50$ & $52.50$ & $88.46$ & $80.77$ & $95.38$ & $79.23$ & $81.95$ & $74.64$ \\
\hline
o1-preview & $93.08$ & {$95.38$} & {$97.69$} & {$96.15$} & {$86.92$} & {{$86.15$}} & {{$90.00$}} & $63.08$ & {$72.50$} & {{$85.00$}} & $88.46$ & {$89.23$} & {$98.46$} & {{$96.15$}} & {$89.59$} & {$87.31$} \\
o1-mini & $90.77$ & $76.92$ & $91.54$ & $88.46$ & {$84.62$} & $68.46$ & {$81.54$} & $56.15$ & $55.83$ & {$70.00$} & $80.00$ & $83.85$ & $91.54$ & {$83.08$} & $82.26$ & $75.27$ \\
\midrule  \multicolumn{17}{c}{zeroshot Input-Output prompt}\\
 \midrule
LLAMA 405B & $88.46$ & $83.08$ & $90.77$ & $90.77$ & $85.38$ & $83.08$ & $84.46$ & $50.00$ & $74.17$ & $72.50$ & $77.69$ & $89.23$ & $83.08$ & $69.23$ & $83.43$ & $76.84$ \\
GPT-4o Mini & $70.77$ & $66.92$ & $68.46$ & $80.77$ & $80.00$ & $58.46$ & $54.62$ & $21.54$ & $57.50$ & $55.83$ & $56.92$ & $44.62$ & $64.62$ & $66.15$ & $64.7$ & $56.33$ \\
GPT-4o & $68.46$ & $83.08$ & $71.54$ & $84.62$ & $74.62$ & $77.69$ & $56.15$ & $37.69$ & $60.00$ & $69.17$ & $59.23$ & $86.92$ & $76.92$ & $80.00$ & $66.7$ & $74.17$ \\
\hline
o1-preview & $88.46$ & {$96.92$} & {$95.38$} & $94.62$ & {$86.15$} & {{$86.15$}} & {$88.46$} & {{$70.77$}} & $69.17$ & {{$88.33$}} & $85.38$ & $86.92$ & {$96.15$} & {$93.85$} & {$87.02$} & {$88.22$} \\
o1-mini & {$90.77$} & {$96.92$} & {$93.08$} & $92.31$ & $83.85$ & $79.23$ & {$83.08$} & $54.62$ & $55.00$ & $60.83$ & $90.00$ & $89.23$ & $95.38$ & {$91.54$} & $84.45$ & {$80.67$} \\
    \end{tabular}
    }
    \caption{Comparison of o1 Reasoning Model with the best performing LLMs on \numoftasks{} ACPBench tasks (Bool as well as MCQ). The right-most column is mean across tasks.}
    \label{tab:compare_o1}
\end{table*}

%% file: appendix.tex
\pagebreak

\section{Appendix}

\subsection{ACPBench Task Examples}

Table~\ref{tab:example_prob} exemplified the domain description, the problem description and the goal description for each domain. For each task we use all or subset of these descriptions as context. Table~\ref{tab:example_tasks} indicates what is included as part of the context for each of the tasks and also provide one example of boolean and multi-choice questions each. 

\subsection{Pretrained LLMs}

Our paper presents domain-wise performance of few selected models in Figure ~\ref{fig:domain_analysis}, where we only presented results for the MCQ due to space constraints. In this section, we present the domain-wise analysis for all the \numofmodels{} pretrained LLMs for both boolean and multi-choice questions in Table~\ref{tab:domain_wise}. 

In the main paper, Table~\ref{tab:compare_models} presents accuracy values with 2-shot COT prompting. We also attempted zeroshot Input Output (IO) prompt, the trends were similar. These zeroshot IO results aggregated over $5$ runs are presented in Table~\ref{tab:compare_models_io}.

\subsection{Finetuned LLM}

 Tables \ref{fig:boolperdomain} and Table \ref{fig:mcqperdomain} show per-domain comparision of the 7 tasks between the Granite (code 8B) Base model and the finetuned model, on multiple choice questions and boolean questions respectively. The ``Diff'' column shows the average gain in performance. Generally we may see a greater performance gain in the seen domains as they have been included in the training set as oppose to the unseen domains. Note, due to memory limitations we were not able to test the Alfworld domain on action reachability. 

In Tables~\ref{tab:finetuned_train} and~\ref{tab:finetuned_test}, we compare te finetuned model against the best performing model (last column). These values are obtained by looking at the aggregated performance on train and test domains respectively. We present performance of all the models on train and test domains in Tables~\ref{tab:compare_models_train} and ~\ref{tab:compare_models_test}.

\subsection{Ablation: Prompt Style}

\begin{figure}[!t]
    \centering
    \includegraphics[width=\columnwidth]{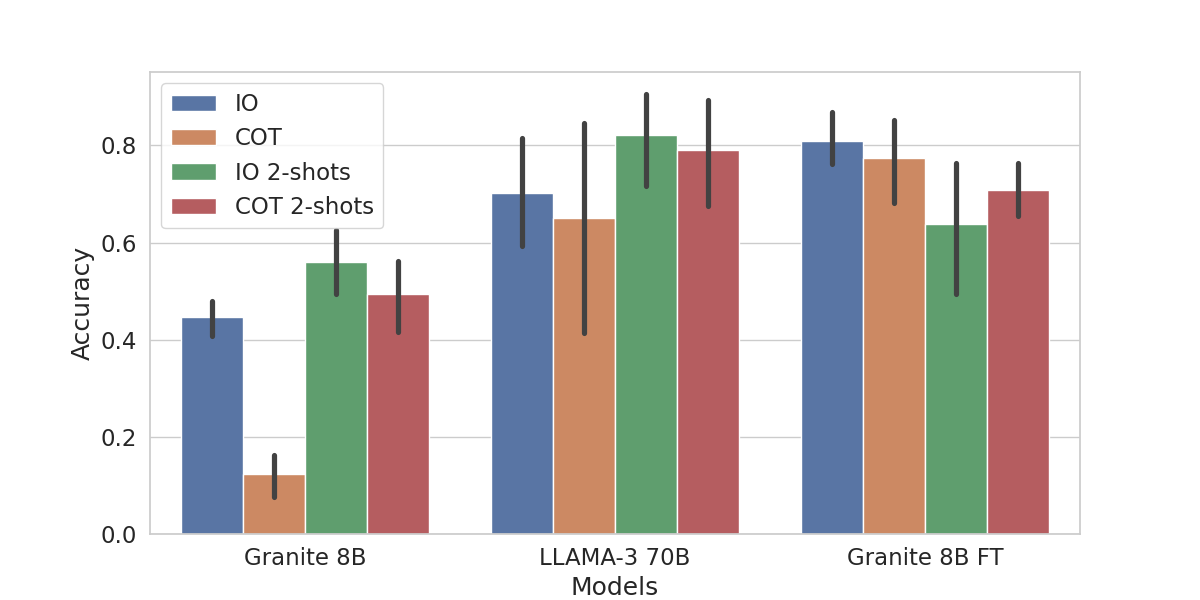}
    \caption{Comparison of different prompt styles on two pretrained models: Granite 8B and LLAMA-3 70B, and finetuned Granite 8B model for boolean tasks in \numoftesting{} testing domains.}
    \label{fig:promptstyle_bool}
\end{figure}

In our paper, we present an ablation to compare prompting styles. For that analysis; we presented results on test domains for multi-choice questions in Figure~\ref{fig:promptstyle}. Here, in Figure~\ref{fig:promptstyle_bool}, we present performance on boolean questions. 

\begin{figure}[!t]
    \centering
    \includegraphics[width=\columnwidth]{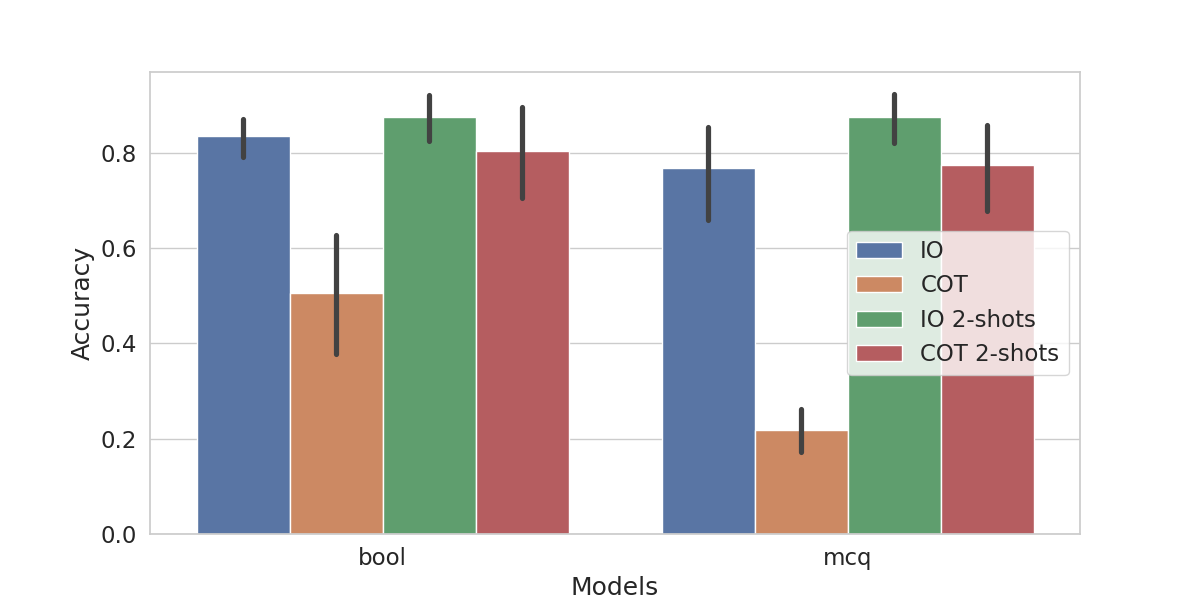}
    \caption{Comparison of different prompt styles on LLAMA 3.1 405B for Bool \& MCQ tasks in all ACPBench domains.}
    \label{fig:promptstyle_llama}
\end{figure}

We also compared the prompt-style on LLAMA 3.1 405B model. Figure~\ref{fig:promptstyle_llama} presents aggregated results on \numoftasks{} ACPBench tasks for all the domains. Although we were only run this experiment once due to resource constraint, we see that COT and IO 2-shot has significant different in performance. The difference is IO 2-shot vs COT 2-shot is significant.

\input{example_problem_goal}

\input{example_tasks}

\input{new_domain_wise}
\input{compare_all_models_IO} %

\input{finetuned-per-domain} %

\input{all_llms_training}
\input{all_llms_testing}

%% file: example_problem_goal.tex
\clearpage
{
\onecolumn
\begin{longtblr}[
  caption = {Example of domain description, problem description and goal description of \numofdomains{} domains in ACPBench.},
  label = {tab:example_prob},
]{
  colspec = {c|X|X|X}, width = \textwidth
}
\hline
   & Domain Desc. & Problem Desc. & Goal Desc.\\
   \hline
  \rotatebox{270}{Blocksworld} & This is a blocksworld domain where blocks can be placed on top of each other or on the table. There is one robotic arm that can move the block. & There are 3 blocks. Currently, the robotic arm is empty. The following block(s) is on the table: block\_1. The following block(s) are stacked on top of another block: block\_3 is on block\_2 and block\_2 is on block\_1. & The goal is to reach a state where the following facts hold: The block block\_3 is currently situated above the block block\_1.\\
\hline
\rotatebox{270}{Logistics} & There are several cities, each containing several locations, some of which are airports. There are also trucks, which can drive within a single city, and airplanes, which can fly between airports. The goal is to get some packages from various locations to various new locations. There are 2 trucks and 1 airplane, as well as 4 packages. There are 6 locations across 2 cities. & The locations are in cities as follows: l0-1, l0-2, and l0-0 are in c0; l1-2, l1-1, and l1-0 are in c1. Currently, t0 and p0 are at l0-0, t1 is at l1-1, p3 and p1 are at l0-2, p2 and a0 are at l1-0.  & The goal is to reach a state where the following facts hold: p2 is at l0-1, p0 is at l0-1, p1 is at l0-1, and p3 is at l0-1. \\
\hline 
\rotatebox{270}{Grippers} & This is a grippers domain, where there is a robot with two grippers. The robot can carry a ball in each. The goal is to take the balls from one room to another. There are 1 robot, 5 rooms, and 4 balls, numbered consecutively. & Currently, the robot robot1 is at room4 and both grippers are free. Additionally, ball3 is at room5, ball4 and ball2 are at room1, ball1 is at room2. & The goal is to reach a state where the following facts hold: Ball ball3 is at room4 location. \\
\hline 

\rotatebox{270}{Grid} &  A robot is in a grid and can only move to places that are connected to its current position. The grid size is 5x5, and the locations are of the form fi-jf (e.g., f3-2f or f0-1f). The grid cells are connected to their neighbors (e.g., f1-2f is connected to the four neighbors f0-2f, f2-2f, f1-1f, and f1-3f). Some positions on the grid are locked and can be opened with a key of a matching shape. The robot has an arm that can pick up a key when the key is in same location as the robot and the arm is empty.  There are 2 keys in 1 different shapes: Key key0-0 is of shape shape0, Key key0-1 is of shape shape0. & Currently, the robot is at position f0-2f and its arm is empty. All the positions are open except the following: f0-1f has shape0 shaped lock, f2-1f has shape0 shaped lock. Key key0-0 is at position f2-2f. Key key0-1 is at position f4-1f. &  The goal is to reach a state where the following facts hold: Key key0-0 is at f3-1f location and Key key0-1 is at f4-1f location. \\
\hline 
\rotatebox{270}{Ferry} & This is a ferry domain, where the task is to transport cars from their start to their goal locations, using a ferry. Each location is accessible by ferry from each other location. The cars can be debarked or boarded, and the ferry can carry only one car at a time. & There are 5 locations and 3 cars, numbered consecutively. Currently, the ferry is at l0 location and it is empty. The cars are at locations as follows: c0 and c2 are at l4; c1 is at l0. & The goal is to reach a state where the following facts hold: Car c1 is at location l3, Car c0 is at location l3, and Car c2 is at location l3. \\
\hline 

\rotatebox{270}{FloorTile} & A set of robots use different colors to paint patterns in floor tiles. The robots can move around the floor tiles in four directions (up, down, left and right). Robots paint with one color at a time, but can change their spray guns to any available color. However, robots can only paint the tile that is in front (up) and behind (down) them, and once a tile has been painted no robot can stand on it. Robots need to paint a grid with black and white, where the cell color is alternated always. There are 2 robots and 12 tiles. The tiles locations are: tile\_6 is to the right of tile\_5, tile\_12 is to the right of tile\_11, tile\_8 is to the right of tile\_7, tile\_5 is to the right of tile\_4, tile\_11 is to the right of tile\_10, tile\_3 is to the right of tile\_2, tile\_9 is to the right of tile\_8, and tile\_2 is to the right of tile\_1. Further, tile\_4 is down from tile\_7, tile\_8 is down from tile\_11, tile\_1 is down from tile\_4, tile\_9 is down from tile\_12, tile\_5 is down from tile\_8, tile\_7 is down from tile\_10, tile\_6 is down from tile\_9, tile\_3 is down from tile\_6, and tile\_2 is down from tile\_5.  &Currently, robot robot1 is at tile\_8 and holding color white and robot robot2 is at tile\_7 and holding color black; tile\_12, tile\_4, tile\_3, tile\_1, tile\_9, tile\_2, tile\_10, tile\_11, tile\_5, and tile\_6 are clear. &  The goal is to reach a state where the following facts hold: Tile tile\_7 is painted in black color, Tile tile\_12 is painted in white color, Tile tile\_4 is painted in white color, Tile tile\_8 is painted in white color, Tile tile\_11 is painted in black color, Tile tile\_6 is painted in white color, Tile tile\_9 is painted in black color, Tile tile\_10 is painted in white color, and Tile tile\_5 is painted in black color.\\
\hline 

\rotatebox{270}{Rovers} & This is a Rovers domain where rovers must navigate between waypoints gathering data and transmitting it back to a lander. Rovers cannot navigate to all waypoints and this makes particular routes impassable to some of the rovers. Data transmission is also constrained by the visibility of the lander from the waypoints. There are 2 rovers, 5 waypoints, 2 stores, 2 cameras, 2 objectives numbered consecutively. Further, there is 1 lander and 3 modes for the camera namely colour, high resolution, and low resolution.  & Rover(s) rover0 and rover1 are equipped for soil analysis. Rover(s) rover1 is equipped for rock analysis. Rover(s) rover0 and rover1 are equipped for imaging. Rover rover0 has store store0. Rover rover1 has store store1. Rover rover0 has camera0 on board. Rover rover1 has camera1 on board. Camera camera1 can be calibrated on objective0. Camera camera0 can be calibrated on objective0. Camera camera1 supports colour and low\_res. Camera camera0 supports colour and low\_res. Rover rover0 can traverse from waypoint4 to waypoint1, waypoint0 to waypoint1, waypoint1 to waypoint0, waypoint1 to waypoint4. Rover rover1 can traverse from waypoint0 to waypoint2, waypoint1 to waypoint2, waypoint2 to waypoint1, waypoint2 to waypoint0. Waypoint(s) are visible from waypoint2: waypoint3, waypoint0, and waypoint1. Waypoint(s) are visible from waypoint1: waypoint4, waypoint2, and waypoint0. Waypoint(s) are visible from waypoint0: waypoint4, waypoint2, waypoint1, and waypoint3. Waypoint(s) are visible from waypoint3: waypoint0 and waypoint2. Waypoint(s) are visible from waypoint4: waypoint0 and waypoint1. Objective objective0 is visible from waypoint1 and waypoint2. Objective objective1 is visible from waypoint4. Lander general is at waypoint waypoint3.  Currently, Rover rover0 is at waypoint0. Rover rover1 is at waypoint2. Rocks can be sampled at the following location(s): waypoint0 and waypoint1. Soil can be sampled at the following location(s): waypoint0. Rovers rover0 and rover1 are available. Store(s) store0 and store1 are empty. & The goal is to reach a state where the following facts hold: Image objective0 was communicated in mode colour, Image objective1 was communicated in mode low\_res, Rock data was communicated from waypoint waypoint0; Rock data was communicated from waypoint waypoint1;, Soil data was communicated from waypoint waypoint0;, and Image objective0 was communicated in mode low\_res.  \\
\hline 

\rotatebox{270}{Visitall} & This is a visitall domain where a robot in a grid must visit all the cells or places in the grid. There are some unavailable places in the grid. The grid size is 4x5, and the location cell names are of the form loc-xi-yj (e.g., loc-x0-y2 or loc-x1-y1). The grid cells are connected to their available neighbors. The unavailable cells are loc-x2-y3, loc-x1-y2, and loc-x0-y4. & Currently, the robot is in place loc-x0-y2.Place loc-x0-y2 has been visited. & The goal is to reach a state where the following facts hold: Place loc-x2-y0 has been visited, Place loc-x3-y3 has been visited, Place loc-x3-y4 has been visited. \\
\hline 
\rotatebox{270}{Depot} & This is a depot domain, a combination of blocks and logistics. In this domain, trucks can transport crates, the crates can be stacked onto pallets using hoists. There are 2 depots, 4 hoists, 4 pallets, 2 distributors, 2 crates, 2 trucks, numbered consecutively. &  Currently, crate1, crate0, pallet1, and pallet2 are clear; hoist1, hoist3, hoist0, and hoist2 are available; pallet0 is at depot0, hoist3 is at distributor1, truck1 is at depot0, pallet2 is at distributor0, hoist2 is at distributor0, truck0 is at distributor0, hoist1 is at depot1, crate1 is at distributor1, crate0 is at depot0, pallet3 is at distributor1, pallet1 is at depot1, and hoist0 is at depot0; crate0 is on pallet0 and crate1 is on pallet3. & The goal is to reach a state where the following facts hold: crate1 is on pallet0 and crate0 is on crate1. \\
\hline 
\rotatebox{270}{Goldminer} & A robotic arm is in a grid and can only move to locations that are connected to its current location. The 3x4 grid locations may have gold, hard rocks, or soft rocks. Rocks cannot be moved. The robotic arm can pick up laser or bomb. Only one item can be picked at a time. There is one laser is the grid that can be used to clear rocks. Robotic arm can fire laser at a location from a connected location. The locations are of the form fi-jf (e.g., f3-2f or f0-1f). The grid cells are connected to their neighbors (e.g., f1-2f is connected to the four neighbors f0-2f, f2-2f, f1-1f, and f1-3f).  If a bomb is picked, it cannot be placed back. It can only be detonated at connected location that have soft rock. Bomb supply is available at f0-0f location.  &  Currently, the robot is at position f2-0f and its arm is empty. The following locations have hard rock: f2-1f, f0-3f, and f2-2f. The following locations have soft rock: f0-2f, f1-2f, f2-3f, f1-3f, f1-1f, and f0-1f. The gold is at f1-3f location. The laser is at f0-0f location. & The goal is to reach a state where the following facts hold: The robot is holding gold. \\
\hline 
\rotatebox{270}{Satellite} & This domain consists of satellite(s) equipped with various instruments that can be switched on when the power is available. Each instrument has a calibration target object and supports taking images of objects in particular modes. When the instrument power is switched on, it is not calibrated. To calibrate an instrument, the satellite should point to the calibration target object and the instrument should be powered on. To take an image of an object, the satellite must point to that object and the instrument must be calibrated. There are 10 satellite(s), numbered consecutively. There are 7 possible target object(s): groundstation1, groundstation0, star2, planet5, star4, planet6, groundstation3. There are 3 image mode(s): image1, thermograph0, infrared2. There are 16 instrument(s), numbered consecutively.  Satellite satellite0 has following instruments onboard: instrument0. $\cdots$ Instrument instrument11 supports image of mode infrared2 and its calibration target is groundstation3. Instrument instrument5 supports image of mode thermograph0 and its calibration target is groundstation1. $\cdots$  &   Currently, Satellite satellite6 is pointing to groundstation1. Satellite satellite5 is pointing to groundstation3. Satellite satellite3 is pointing to groundstation1. Satellite satellite2 is pointing to planet6. Satellite satellite1 is pointing to groundstation0. Satellite satellite7 is pointing to star2. Satellite satellite0 is pointing to groundstation1. Satellite satellite9 is pointing to star4. Satellite satellite4 is pointing to planet6. Satellite satellite8 is pointing to star2. Power is available on the following satellite(s): satellite1, satellite2, satellite5, satellite0, satellite6, satellite7, satellite8, satellite9, satellite4, satellite3. & The goal is to reach a state where the following facts hold: A thermograph0 mode image of target planet5 is available, Satellite satellite6 is pointing to star4, A infrared2 mode image of target planet6 is available, and Satellite satellite8 is pointing to planet6. \\
\hline 
\rotatebox{270}{Swap} & This is a swap domain where agents are swapping items or roles. Each agent is always assigned a single item/role. The goal is to obtain desired items/roles assigned. There are 8 agents: vic, alice, zoe, dave, heidi, carol, michelle, and xena. There are 8 items/roles: necklace, whale, iceskates, frisbee, guitar, quadcopter, slinky, and zebra. &  Currently, zoe is assigned frisbee, heidi is assigned necklace, carol is assigned guitar, michelle is assigned zebra, dave is assigned slinky, xena is assigned whale, alice is assigned iceskates, and vic is assigned quadcopter. & The goal is to reach a state where the following facts hold: heidi is assigned guitar, michelle is assigned quadcopter. \\
\hline 

\rotatebox{270}{Alfworld} &  This is an alfworld domain where an agent is asked to carry different tasks such as: picking up objects, opening or closing receptacles, warming up an object in a microwave, cleaning an object in a sink, or toggling an object.  There are 21 object types: 3 alarmclocks, 1 baseballbat, 1 basketball, 2 blindss, 1 book, 3 bowls, 3 cds, 3 cellphones, 2 chairs, 1 creditcard, 1 desklamp, 2 keychains, 2 laptops, 1 laundryhamperlid, 1 lightswitch, 1 mirror, 2 mugs, 3 pencils, 1 pen, 2 pillows, 2 windows, 7 receptacle types: 1 bed, 2 desks, 6 drawers, 1 garbagecan, 1 laundryhamper, 1 safe, 6 shelves, and 27 locations all numbered consecutively.  The receptacles are at locations as follows. laundryhamper1 is at location8. shelf1 is at location20. drawer1 is at location21. bed1 is at location13. shelf3 is at location11. shelf4 is at location23. desk2 is at location10. drawer5 and drawer4 are at location12. desk1 is at location3. drawer6 is at location1. safe1 is at location6. shelf2 is at location25. shelf6 is at location24. drawer3 is at location17. drawer2 is at location18. shelf5 is at location22. garbagecan1 is at location2. &  Currently, the objects are at locations as follows. bowl1, alarmclock1, mug1, cd1, and pencil1 are at location3. window2 is at location4. basketball1 is at location7. pen1, mug2, pencil3, cellphone2, and cd3 are at location10. pillow1, laptop2, book1, cellphone1, laptop1, and pillow2 are at location13. chair1 is at location21. laundryhamperlid1 is at location8. baseballbat1 is at location9. pencil2 and creditcard1 are at location22. desklamp1, bowl2, and alarmclock3 are at location23. bowl3 is at location24. keychain2 and keychain1 are at location6. mirror1 is at location19. cd2 is at location2. lightswitch1 is at location14. cellphone3 is at location12. chair2 is at location26. blinds2 is at location15. blinds1 is at location16. alarmclock2 is at location11. window1 is at location5. agent agent1 is at location location27. The objects are in/on receptacle as follows. pen1, cellphone2, cd3, bowl2, mug2, desklamp1, pencil3, and alarmclock3 are on desk2. bowl1, mug1, alarmclock1, pencil1, and cd1 are on desk1. pencil2 and creditcard1 are on shelf5. keychain2 and keychain1 are in safe1. cd2 is in garbagecan1. laptop2, laptop1, book1, cellphone1, pillow2, and pillow1 are in bed1. cellphone3 is in drawer5. alarmclock3, bowl2, and desklamp1 are on shelf4. alarmclock2 is on shelf3. bowl3 is on shelf6. drawer3, drawer6, safe1, and drawer1 are closed. desklamp1 is off. Nothing has been validated. agent1's hands are empty. &  The goal is to reach a state where the following facts hold: an object of type book is examined under an object of type desklamp.\\
\hline 
\end{longtblr}
}
\twocolumn

%% file: example_tasks.tex
\clearpage
{
\onecolumn
\begin{longtblr}[
  caption = {Example questions for \numoftasks{} ACPBench tasks. For each question, an LLM was provided with context and then the question. The context contains natural language descriptions.},
  label = {tab:example_tasks},
]{
  colspec = {c|c|X|X}, width = \textwidth
}
\hline
         Task & Context & Bool Questions & MCQ Questions \\
         \midrule
\rotatebox{270}{Applicability} & {domain \\ +\\ problem } &
        Is the following action applicable in this state:  \texttt{debark the car c2 from the ferry to location l1?} &
        Which of the following actions will be applicable in this state?     \newline A. \texttt{travel by sea from location l2 to location l1.} 
        \newline
        B. \texttt{debark the car c2 to location l0 from the ferry.} 
        \newline
        C. \texttt{travel by sea from location l0 to location l1.}
        \newline 
        D. \texttt{board the car c5 at location l1 on to the ferry.}\\
        \midrule
\rotatebox{270}{Progression} & {domain \\ +  \\ problem} &
        Will the fact \texttt{"The ferry is empty"} hold after performing the action \texttt{"embark the car c0 at location l1 on to the ferry"} in the current state? &  
        Which of the following facts hold after performing the action \texttt{"sail from location l1 to location l0"} in the current state? \newline 
        A. \texttt{The ferry is at l0 location and Car c3 is at location l0.}
        \newline
        B. \texttt{The ferry is at l0 location and The ferry is at l1 location.}
        \newline
        C. \texttt{The ferry is at l0 location and Car c4 is at location l0.}
        \newline
        D. \texttt{Car c4 is at location l0 and Car c3 is at location l0.} \\
        \midrule
\rotatebox{270}{Reachability} & {domain \\ +  \\ problem}
& Is it possible to transition to a state where the following holds: \newline
        \texttt{The ferry is at l0 location and Car c0 is on board the ferry?} & Which of the following options can hold in a state that can potentially be reached? \newline 
        A. \texttt{The ferry is at c9 location and Car c6 is at location l2.}  \newline
        B. \texttt{The ferry is at l0 location and The ferry is at l2 location.} \newline
        C. \texttt{Car l2 is on the ferry and Car c4 is at location l0.} \newline
        D. \texttt{There are no cars on the ferry and Car c5 is at location l1.} \\
        \midrule 
\rotatebox{270}{Validation} & {domain \\ +  \\ problem \\ + \\ goal}& Is the following sequence of actions a plan for the current state? \newline
\texttt{sail from location l2 to location l0}, \newline \texttt{board car c6 at location l0}, \newline 
\texttt{sail from location l0 to location l2}, ...
        &
        Which of the following claims is true with regard to the following sequence of actions \newline
        \texttt{board the car c12 at the location l1}, \newline 
        \texttt{travel by sea from location l1 to location l0},\newline 
        \texttt{debark the car c12 from the ferry to location l0}, \newline 
        \texttt{board the car c33 at the location l0}, ... \newline 
        A. \texttt{The sequence is not valid.}\newline 
        B. \texttt{The sequence is applicable, but does not achieve the goal.}\newline  
        C. \texttt{The sequence is a plan.}\newline 
        D. \texttt{The sequence is not applicable.}\\
        \midrule 
\rotatebox{270}{Act. Reachability} & {domain \\ + \\ problem} & 
Is it possible to transition to a state where the action \texttt{"embark the car l2 at location c8 on to the ferry"} can be applied? & Which of the following actions can eventually be applied?  \newline  
A. \texttt{board the car c20 at location l0.} \newline
B. \texttt{travel by sea from location c43 to location c4.} \newline  
C. \texttt{debark the car c2 to location c8 from the ferry.} \newline
D. \texttt{board the car c26 at location c23.} \\
\midrule
\rotatebox{270}{Justification} & {domain \\ + \\ problem \\ + \\ goal}  & Given the plan: \newline 
\texttt{"board the car c13 at location l1 on to the ferry, sail from location l1 to location l3, debark car c13 to location l3 from the ferry, board the car c29 at location l3 on to the ferry,..."}; \newline can the following action be removed from this plan and still have a valid plan: \texttt{sail from location l1 to location l3}? &   Given the plan: \newline 
\texttt{"travel by sea from location l2 to location l1, board the car c0 at the location l1, travel by sea from location l1 to location l0, debark the car c0 from the ferry to location l0,..."}; \newline
which of the following pairs of consecutive actions can be removed from this plan and still have a valid plan? \newline 
A. \texttt{board the car c0 at the location l1 and travel by sea from location l1 to location l0.} \newline
B. \texttt{board the car c0 at the location l0 and travel by sea from location l0 to location l2.} \newline  
C. \texttt{board the car c6 at the location l0 and travel by sea from location l0 to location l2.} \newline
D. \texttt{debark the car c0 from the ferry to location l0 and board the car c0 at the location l0?} 
\\
\hline
\rotatebox{270}{Landmark} & {domain \\ + \\ problem \\ + \\ goal} & Is the following fact a landmark (must hold at some point along any plan) for the current state? \newline 
\texttt{There are no cars on the ferry} &  Which of the following facts is a landmark (must hold at some point along any plan) for the current state? \newline 
A. \texttt{Car c7 is on board the ferry.} \newline
B. \texttt{Car c5 is at location l2.} \newline
C. \texttt{Car c7 is at location l2.} \newline
D. \texttt{Car c0 is at location l1.}\\
\hline
\end{longtblr}
}
\twocolumn

%% file: new_domain_wise.tex
\begin{table*}[t]
    \centering
    \resizebox{\textwidth}{!}{
        \begin{tabular}{l|cc|cc|cc|cc|cc|cc|cc|cc|cc|cc|cc|cc|cc|}
        Model & \multicolumn{2}{c|}{Blocksworld} & \multicolumn{2}{c|}{Logistics} & \multicolumn{2}{c|}{Grippers} & \multicolumn{2}{c|}{Grid} & \multicolumn{2}{c|}{Ferry} & \multicolumn{2}{c|}{FloorTile} & \multicolumn{2}{c|}{Rovers} & \multicolumn{2}{c|}{VisitAll} & \multicolumn{2}{c|}{Depot} & \multicolumn{2}{c|}{Goldminer} & \multicolumn{2}{c|}{Satellite} & \multicolumn{2}{c|}{Swap} & \multicolumn{2}{c|}{Alfworld} \\
         & Bool & MCQ & Bool & MCQ& Bool & MCQ& Bool & MCQ& Bool & MCQ& Bool & MCQ& Bool & MCQ& Bool & MCQ& Bool & MCQ& Bool & MCQ& Bool & MCQ& Bool & MCQ& Bool & MCQ\\
        \midrule
Phi-3 128K  & $40.0$ & $27.14$ & $61.43$ & $25.71$ & $60.0$ & $28.57$ & $50.0$ & $45.71$ & $55.71$ & $32.86$ & $45.71$ & $31.43$ & $48.57$ & $27.14$ & $71.43$ & $45.71$ & $67.14$ & $44.29$ & $58.57$ & $32.86$ & $50.0$ & $42.86$ & $55.71$ & $31.43$ & $51.43$ & $35.71$  \\
Gemma 7B  & $44.29$ & $27.43$ & $50.0$ & $24.86$ & $51.71$ & $22.29$ & $50.86$ & $33.71$ & $54.29$ & $28.29$ & $54.0$ & $33.71$ & $57.14$ & $26.57$ & $62.57$ & $30.57$ & $40.57$ & $25.43$ & $42.29$ & $34.86$ & $50.86$ & $31.14$ & $53.14$ & $28.0$ & $52.86$ & $27.14$  \\
Granite 7B  & $39.71$ & $25.43$ & $45.71$ & $27.14$ & $52.86$ & $22.86$ & $51.43$ & $30.0$ & $52.86$ & $32.0$ & $40.0$ & $24.29$ & $47.14$ & $30.0$ & $64.29$ & $38.57$ & $50.0$ & $38.57$ & $42.86$ & $25.71$ & $37.14$ & $34.29$ & $58.29$ & $32.86$ & $27.14$ & $27.14$  \\
Mistral 7B  & $40.31$ & $18.46$ & $56.15$ & $26.92$ & $61.54$ & $25.23$ & $50.77$ & $30.77$ & $53.08$ & $31.54$ & $64.0$ & $23.08$ & $62.31$ & $32.15$ & $60.77$ & $38.77$ & $54.92$ & $35.08$ & $54.62$ & $16.15$ & $49.23$ & $25.38$ & $54.62$ & $31.54$ & $46.15$ & $36.15$  \\
Mistral instruct 7B  & $47.14$ & $36.86$ & $53.71$ & $36.29$ & $64.29$ & $54.29$ & $57.71$ & $38.0$ & $65.71$ & $32.57$ & $54.29$ & $30.0$ & $43.71$ & $28.57$ & $59.71$ & $35.71$ & $63.14$ & $38.57$ & $62.86$ & $41.43$ & $43.14$ & $32.86$ & $52.86$ & $44.86$ & $40.0$ & $31.43$  \\
Granite-code 8B  & $55.71$ & $15.71$ & $58.57$ & $39.43$ & $67.43$ & $22.86$ & $41.43$ & $24.0$ & $50.86$ & $33.14$ & $44.29$ & $19.43$ & $52.29$ & $27.43$ & $62.86$ & $38.57$ & $42.0$ & $28.29$ & $50.0$ & $28.57$ & $47.14$ & $34.29$ & $54.29$ & $31.71$ & $47.14$ & $31.43$  \\
Granite8b code instruct  & $48.57$ & $31.43$ & $52.86$ & $33.14$ & $55.71$ & $32.86$ & $47.14$ & $30.57$ & $57.14$ & $25.71$ & $41.43$ & $21.43$ & $51.14$ & $34.29$ & $58.57$ & $36.29$ & $47.14$ & $31.43$ & $40.0$ & $32.86$ & $49.43$ & $38.57$ & $51.43$ & $41.43$ & $50.29$ & $28.57$  \\
LLAMA-3 8B  & $54.29$ & $44.29$ & $61.43$ & $48.57$ & $67.14$ & $40.0$ & $58.29$ & $47.43$ & $72.86$ & $31.43$ & $47.14$ & $45.71$ & $52.86$ & $34.57$ & $71.43$ & $55.71$ & $60.86$ & $36.86$ & $58.86$ & $50.0$ & $65.71$ & $51.43$ & $64.29$ & $40.0$ & $55.71$ & $41.43$  \\
LLAMA-3.1 8B  & $54.29$ & $47.14$ & $52.86$ & $35.71$ & $58.57$ & $37.14$ & $52.86$ & $38.57$ & $48.57$ & $40.0$ & $51.43$ & $40.0$ & $60.0$ & $30.0$ & $51.43$ & $52.86$ & $51.43$ & $48.57$ & $48.57$ & $42.86$ & $41.43$ & $34.29$ & $58.57$ & $42.86$ & $42.86$ & $44.29$  \\
Mixtral 8x7B  & $65.43$ & $48.29$ & $60.29$ & $40.0$ & $71.43$ & $46.0$ & $65.71$ & $59.71$ & $80.0$ & $47.14$ & $53.71$ & $46.0$ & $53.43$ & $38.57$ & $81.71$ & $60.0$ & $71.14$ & $56.0$ & $65.14$ & $49.71$ & $64.0$ & $65.43$ & $63.71$ & $51.43$ & $49.71$ & $52.86$  \\
Granite 13B  & $46.29$ & $26.57$ & $64.29$ & $26.86$ & $57.43$ & $27.14$ & $37.14$ & $37.14$ & $52.86$ & $26.0$ & $40.86$ & $24.29$ & $49.43$ & $12.29$ & $44.86$ & $31.71$ & $38.0$ & $35.71$ & $40.0$ & $13.71$ & $51.43$ & $21.43$ & $44.29$ & $31.43$ & $48.0$ & $28.0$  \\
Codestral 22B  & $71.43$ & $31.43$ & $70.0$ & $38.57$ & $80.0$ & $35.71$ & $68.57$ & $42.86$ & $68.57$ & $45.71$ & $60.0$ & $32.86$ & $64.29$ & $35.71$ & $60.0$ & $48.57$ & $68.57$ & $38.57$ & $64.29$ & $41.43$ & $65.71$ & $42.86$ & $62.86$ & $52.86$ & $60.0$ & $40.0$  \\
Mixtral 8x22B  & $46.86$ & $44.29$ & $65.14$ & $36.29$ & $66.57$ & $38.57$ & $50.57$ & $45.14$ & $67.71$ & $34.86$ & $39.71$ & $22.0$ & $45.71$ & $32.0$ & $60.57$ & $42.57$ & $69.43$ & $42.86$ & $50.0$ & $43.43$ & $61.71$ & $44.86$ & $66.57$ & $44.86$ & $20.0$ & $24.86$  \\
Deepseek-33b instruct  & $51.43$ & $31.14$ & $64.29$ & $42.86$ & $62.86$ & $20.86$ & $56.0$ & $45.71$ & $61.43$ & $30.0$ & $48.57$ & $32.86$ & $60.0$ & $35.71$ & $70.0$ & $44.29$ & $60.0$ & $35.71$ & $57.14$ & $36.29$ & $54.29$ & $38.57$ & $46.86$ & $22.86$ & $48.57$ & $35.71$  \\
CodeLLAMA 34B  & $53.14$ & $42.0$ & $58.57$ & $34.29$ & $58.57$ & $27.14$ & $52.86$ & $37.14$ & $61.43$ & $31.43$ & $58.57$ & $35.71$ & $60.0$ & $21.43$ & $62.86$ & $54.29$ & $52.86$ & $36.86$ & $57.14$ & $37.14$ & $61.14$ & $28.57$ & $61.43$ & $42.86$ & $60.0$ & $30.0$  \\
\midrule
LLAMA-2 7OB  & $51.43$ & $28.57$ & $59.71$ & $24.29$ & $53.43$ & $30.0$ & $50.0$ & $25.71$ & $58.57$ & $14.57$ & $57.14$ & $15.71$ & $53.14$ & $32.86$ & $61.43$ & $42.86$ & $48.57$ & $36.86$ & $57.14$ & $28.57$ & $57.43$ & $31.43$ & $58.57$ & $36.0$ & $47.14$ & $35.71$  \\
CodeLLAMA 70B  & $51.14$ & $27.43$ & $52.86$ & $30.0$ & $60.0$ & $24.57$ & $45.43$ & $42.86$ & $59.71$ & $36.57$ & $46.86$ & $30.29$ & $48.0$ & $32.86$ & $63.71$ & $35.71$ & $46.86$ & $30.0$ & $45.43$ & $40.86$ & $58.86$ & $30.86$ & $45.71$ & $32.57$ & $21.71$ & $34.29$  \\
LLAMA-3 70B  & $81.43$ & $70.0$ & $77.71$ & $80.0$ & $80.0$ & $90.86$ & $84.57$ & $72.86$ & $83.71$ & $78.57$ & $74.29$ & $65.71$ & $75.71$ & $66.57$ & $68.57$ & $72.86$ & $75.71$ & $84.29$ & $82.86$ & $71.43$ & $81.71$ & $58.57$ & $75.71$ & $77.14$ & $72.86$ & $68.86$  \\
LLAMA-3.1 70B  & $74.29$ & $55.43$ & $73.71$ & $69.14$ & $73.14$ & $79.43$ & $60.29$ & $68.0$ & $61.14$ & $78.57$ & $60.29$ & $61.71$ & $67.14$ & $56.57$ & $59.71$ & $62.86$ & $63.14$ & $66.57$ & $70.86$ & $72.57$ & $62.0$ & $53.43$ & $64.86$ & $76.86$ & $64.0$ & $60.29$  \\
LLAMA-3.1 405B  & $81.43$ & $71.43$ & $92.86$ & $85.71$ & $88.57$ & $93.14$ & $83.43$ & $74.29$ & $81.43$ & $81.43$ & $57.14$ & $61.43$ & $80.0$ & $71.43$ & $81.43$ & $82.86$ & $85.71$ & $71.43$ & $81.43$ & $78.29$ & $71.43$ & $72.57$ & $81.43$ & $85.71$ & $70.0$ & $67.14$  \\
\midrule
GPT-4o Mini  & $70.0$ & $38.57$ & $78.57$ & $58.57$ & $77.14$ & $48.57$ & $80.0$ & $58.57$ & $81.43$ & $57.14$ & $64.29$ & $58.57$ & $71.43$ & $32.86$ & $72.86$ & $60.0$ & $75.71$ & $67.14$ & $88.57$ & $64.29$ & $80.0$ & $55.71$ & $84.29$ & $68.57$ & $77.14$ & $67.14$  \\
GPT-4o  & $87.14$ & $75.71$ & $80.0$ & $70.0$ & $80.0$ & $84.29$ & $88.57$ & $82.86$ & $90.0$ & $80.0$ & $61.43$ & $54.29$ & $72.86$ & $55.71$ & $87.14$ & $74.29$ & $77.14$ & $75.71$ & $90.0$ & $80.0$ & $75.71$ & $72.86$ & $94.29$ & $82.86$ & $72.86$ & $74.29$  \\
    \end{tabular}
    }
    \caption{Accuracy values for \numofmodels{} pretrained LLMs on Bool and MCQ tasks in ACPBench; segregated per domain. }
    \label{tab:domain_wise}
\end{table*}

%% file: compare_all_models_IO.tex
\begin{table*}[t]
    \centering
    \resizebox{\textwidth}{!}{
    \begin{tabular}{l@{}|r@{\hspace{0.6\tabcolsep}}r|r@{\hspace{0.6\tabcolsep}}r|r@{\hspace{0.6\tabcolsep}}r|r@{\hspace{0.6\tabcolsep}}r|r@{\hspace{0.6\tabcolsep}}r|r@{\hspace{0.6\tabcolsep}}r|r@{\hspace{0.6\tabcolsep}}r|r@{\hspace{0.6\tabcolsep}}r}
    \multirow{2}{*}{\textbf{Model}}  & \multicolumn{2}{c|}{Applicability} & \multicolumn{2}{c|}{{Progression}}&  \multicolumn{2}{c|}{Reachability} & \multicolumn{2}{c|}{Validation} & \multicolumn{2}{c|}{Action Reach.} &  \multicolumn{2}{c|}{Justification} & \multicolumn{2}{c|}{Landmark} & \multicolumn{2}{c}{Mean}\\
 & Bool & MCQ & Bool & MCQ& Bool & MCQ& Bool & MCQ& Bool & MCQ& Bool & MCQ& Bool & MCQ& Bool & MCQ\\
 \midrule
Phi-3 128K & $66.92$ & $40.77$ & $43.85$ & $56.92$ & $58.46$ & $23.85$ & $55.38$ & $25.38$ & $63.33$ & $27.50$ & $51.54$ & $26.15$ & $73.85$ & $48.46$ & $55.64$ & $35.08$ \\
Gemma 7B & $47.69$ & $25.23$ & $48.62$ & $34.46$ & $53.85$ & $20.00$ & $55.38$ & $29.23$ & $58.17$ & $21.00$ & $48.31$ & $29.08$ & $46.92$ & $33.85$ & $51.73$ & $29.15$ \\
Granite 7B & $49.23$ & $23.85$ & $52.62$ & $35.38$ & $57.38$ & $26.92$ & $55.38$ & $30.00$ & $51.67$ & $20.00$ & $48.00$ & $26.92$ & $52.31$ & $32.31$ & $47.41$ & $30.22$ \\
Mistral 7B & $54.62$ & $28.46$ & $53.08$ & $20.00$ & $49.23$ & $26.15$ & $46.92$ & $22.62$ & $45.00$ & $16.67$ & $49.23$ & $22.31$ & $47.69$ & $29.23$ & $54.91$ & $28.45$ \\
Mistral instruct 7B & $67.69$ & $27.69$ & $50.77$ & $39.23$ & $65.38$ & $23.08$ & $64.15$ & $33.85$ & $53.33$ & $24.17$ & $53.08$ & $25.38$ & $76.15$ & $50.77$ & $54.99$ & $37.41$ \\
Granite-code 8B & $52.31$ & $14.62$ & $51.54$ & $24.62$ & $46.92$ & $13.08$ & $45.38$ & $33.69$ & $41.67$ & $11.17$ & $50.77$ & $16.15$ & $53.85$ & $20.00$ & $52.48$ & $29.12$ \\
Granite8b code instruct & $53.08$ & $13.85$ & $55.69$ & $25.38$ & $49.23$ & $23.08$ & $49.23$ & $34.92$ & $53.33$ & $12.50$ & $50.77$ & $18.46$ & $59.23$ & $28.46$ & $50.53$ & $32.63$ \\
LLAMA-3 8B & $59.23$ & $30.77$ & $51.54$ & $48.46$ & $55.38$ & $25.38$ & $53.08$ & $36.92$ & $55.83$ & $21.67$ & $54.77$ & $26.15$ & $66.15$ & $40.77$ & $61.53$ & $44.05$ \\
LLAMA-3.1 8B & $53.85$ & $36.92$ & $48.46$ & $45.38$ & $52.31$ & $33.85$ & $56.15$ & $28.46$ & $58.33$ & $33.33$ & $49.23$ & $22.31$ & $47.69$ & $45.38$ & $52.23$ & $41.41$ \\
Mixtral 8x7B & $67.08$ & $45.38$ & $52.62$ & $62.15$ & $60.62$ & $49.23$ & $50.62$ & $43.69$ & $44.83$ & $43.83$ & $52.77$ & $24.62$ & $77.23$ & $66.92$ & $65.61$ & $51.46$ \\
Granite 13B & $57.69$ & $36.92$ & $50.00$ & $38.46$ & $58.46$ & $29.23$ & $70.00$ & $38.46$ & $61.67$ & $43.33$ & $68.31$ & $26.92$ & $62.31$ & $60.00$ & $47.79$ & $26.62$ \\
Codestral 22B & $72.31$ & $30.77$ & $56.92$ & $53.85$ & $65.38$ & $24.62$ & $48.46$ & $26.15$ & $54.17$ & $23.33$ & $51.54$ & $23.08$ & $71.54$ & $34.62$ & $67.07$ & $40.97$ \\
Mixtral 8x22B & $50.15$ & $40.92$ & $52.77$ & $47.69$ & $65.38$ & $30.92$ & $56.92$ & $38.46$ & $50.50$ & $28.33$ & $47.69$ & $33.85$ & $64.00$ & $63.08$ & $55.30$ & $38.50$ \\
Deepseek-33b instruct & $60.00$ & $18.92$ & $46.00$ & $29.23$ & $63.08$ & $24.62$ & $56.15$ & $34.31$ & $48.83$ & $15.83$ & $50.77$ & $19.23$ & $70.46$ & $22.31$ & $57.58$ & $35.11$ \\
CodeLLAMA 34B & $60.92$ & $16.92$ & $53.08$ & $42.31$ & $52.31$ & $20.00$ & $48.46$ & $16.92$ & $44.17$ & $14.00$ & $50.77$ & $22.31$ & $69.23$ & $40.77$ & $58.94$ & $35.66$ \\
\midrule
LLAMA-2 7OB & $53.38$ & $27.69$ & $49.38$ & $47.08$ & $38.46$ & $25.38$ & $44.00$ & $36.31$ & $43.17$ & $18.83$ & $51.08$ & $25.85$ & $53.85$ & $40.77$ & $55.57$ & $29.71$ \\
CodeLLAMA 70B & $57.54$ & $32.92$ & $51.08$ & $38.46$ & $46.92$ & $35.23$ & $44.62$ & $34.62$ & $42.50$ & $31.00$ & $50.00$ & $24.62$ & $64.62$ & $46.92$ & $50.26$ & $33.30$ \\
LLAMA-3 70B & $90.00$ & $71.54$ & $62.31$ & $86.92$ & $76.15$ & $78.46$ & $60.15$ & $48.00$ & $54.17$ & $57.67$ & $54.62$ & $83.08$ & $84.62$ & $74.31$ & $78.73$ & $74.36$ \\
LLAMA-3.1 70B & $40.00$ & $72.00$ & $40.31$ & $85.38$ & $65.85$ & $71.69$ & $48.00$ & $40.77$ & $57.00$ & $58.83$ & $35.54$ & $70.92$ & $17.38$ & $76.92$ & $66.43$ & $66.90$ \\
LLAMA-3.1 405B & $88.46$ & $83.08$ & $90.77$ & $90.77$ & $85.38$ & $83.08$ & $84.46$ & $50.00$ & $74.17$ & $72.50$ & $77.69$ & $89.23$ & $83.08$ & $69.23$ & $80.43$ & $77.39$ \\
\midrule
GPT-4o Mini & $70.77$ & $66.92$ & $68.46$ & $80.77$ & $80.00$ & $58.46$ & $54.62$ & $21.54$ & $57.50$ & $55.83$ & $56.92$ & $44.62$ & $64.62$ & $66.15$ & $77.63$ & $56.83$ \\
GPT-4o & $68.46$ & $83.08$ & $71.54$ & $84.62$ & $74.62$ & $77.69$ & $56.15$ & $37.69$ & $60.00$ & $69.17$ & $59.23$ & $86.92$ & $76.92$ & $80.00$ & $81.95$ & $74.64$ \\
    \end{tabular}
    }
    \caption{Accuracy of \numofmodels{} LLMs on \numoftasks{} ACPBench tasks (Bool as well as MCQ) with zeroshot IO prompt. The right-most column is mean across tasks.}
    \label{tab:compare_models_io}
\end{table*}

%% file: finetuned-per-domain.tex
\newcommand{\finetuned}{Finetuned} 
\newcommand{\base}{Base} 

\addtolength{\tabcolsep}{+0.4em}

\begin{table*}[tb]
{\small
 \resizebox{\textwidth}{!}{
\begin{tabular}{l|cc|cc|cc|cc|cc|cc|cc|c}

 & \multicolumn{2}{|c|}{Applicability} & \multicolumn{2}{|c|}{Progression}   & \multicolumn{2}{|c|}{Reachability} &   \multicolumn{2}{|c|}{Validation} &  \multicolumn{2}{|c|}{Action Reach.} &  \multicolumn{2}{|c|}{Justification}  &  \multicolumn{2}{|c|}{Landmark} &  \\ 
Domain & \base & \finetuned & \base & \finetuned & \base & \finetuned & \base & \finetuned & \base & \finetuned & \base & \finetuned & \base & \finetuned & Diff\\ 
\hline

Ferry & 60 & 100 & 30 & 100 & 80 & 100 & 70 & 100 & 50 & 100 & 40 & 100 & 50 & 80 & 42.86\\
Logistics & 50 & 100 & 80 & 100 & 30 & 100 & 30 & 100 & 30 & 100 & 40 & 100 & 50 & 70 & 51.43\\
Blocksworld & 70 & 100 & 80 & 90 & 50 & 100 & 30 & 100 & 70 & 100 & 80 & 100 & 60 & 90 & 34.29\\
Grid & 20 & 100 & 70 & 100 & 80 & 100 & 40 & 100 & 60 & 100 & 60 & 100 & 60 & 60 & 38.57\\
Floortile & 50 & 100 & 50 & 100 & 50 & 100 & 40 & 100 & 30 & 100 & 90 & 90 & 60 & 90 & 44.29\\
Grippers & 60 & 100 & 30 & 100 & 30 & 100 & 60 & 100 & 40 & 100 & 50 & 100 & 70 & 90 & 50.00\\
Rovers & 60 & 90 & 50 & 90 & 30 & 80 & 30 & 100 & 50 & 80 & 70 & 90 & 60 & 80 & 37.14\\
Visitall & 60 & 100 & 30 & 100 & 30 & 100 & 60 & 100 & 30 & 100 & 20 & 100 & 70 & 90 & 55.71\\
\hline
Depot & 50 & 90 & 30 & 70 & 30 & 70 & 30 & 90 & 40 & 80 & 20 & 100 & 30 & 60 & 47.14\\
Goldminer & 70 & 60 & 40 & 90 & 80 & 90 & 30 & 80 & 30 & 50 & 40 & 100 & 60 & 90 & 30.00\\
Satellite & 70 & 40 & 80 & 100 & 50 & 80 & 50 & 80 & 30 & 100 & 50 & 100 & 30 & 70 & 30.00\\
Swap & 20 & 90 & 30 & 90 & 40 & 90 & 70 & 60 & 40 & 100 & 50 & 90 & 70 & 100 & 42.86\\
Alfworld & 40 & 90 & 70 & 50 & 30 & 80 & 50 & 70 & NA & NA & 50 & 100 & 30 & 40 & 26.67\\
\hline
Mean & 52 & 89 & 52 & 91 & 47 & 92 & 45 & 91 & 45 & 93 & 51 & 98 & 54 & 78 & 40.84

\end{tabular}
}
}
\caption{\footnotesize Per-domain comparison of 7 tasks on the boolean questions between the Base model, Granite8b code base, and the Finetuned model. The first 8 domains are domains that are in the training set (seen domains), and the last 5 domains are unseen domains. ``Diff'' shows the average difference between the base and fine-tuned model. }
\label{fig:boolperdomain}
\end{table*}

\begin{table*}[tb]
{\small
 \resizebox{\textwidth}{!}{
\begin{tabular}{l|cc|cc|cc|cc|cc|cc|cc|c}

 & \multicolumn{2}{|c|}{Applicability} & \multicolumn{2}{|c|}{Progression}   & \multicolumn{2}{|c|}{Reachability} &   \multicolumn{2}{|c|}{Validation} &  \multicolumn{2}{|c|}{Action Reach.} &  \multicolumn{2}{|c|}{Justification}  &  \multicolumn{2}{|c|}{Landmark} &  \\ 
Domain & \base & \finetuned & \base & \finetuned & \base & \finetuned & \base & \finetuned & \base & \finetuned & \base & \finetuned & \base & \finetuned & Diff\\ 
\hline

Ferry & 30 & 90 & 40 & 100 & 0 & 100 & 20 & 80 & 10 & 100 & 0 & 90 & 20 & 100 & 77.14\\
Logistics & 10 & 90 & 20 & 100 & 0 & 100 & 40 & 80 & 18 & 100 & 0 & 90 & 10 & 80 & 77.43\\
Blocksworld & 10 & 70 & 20 & 50 & 40 & 90 & 30 & 70 & 10 & 90 & 40 & 100 & 50 & 100 & 52.86\\
Grid & 0 & 100 & 10 & 100 & 20 & 100 & 20 & 100 & 10 & 100 & 40 & 100 & 20 & 100 & 82.86\\
Floortile & 40 & 100 & 50 & 100 & 10 & 100 & 30 & 90 & 20 & 90 & 10 & 90 & 10 & 100 & 71.43\\
Grippers & 20 & 100 & 10 & 100 & 30 & 100 & 70 & 100 & 26 & 100 & 10 & 100 & 10 & 100 & 74.86\\
Rovers & 0 & 90 & 20 & 100 & 0 & 100 & 68 & 80 & 0 & 80 & 0 & 100 & 10 & 60 & 73.14\\
Visitall & 10 & 100 & 10 & 100 & 30 & 100 & 30 & 100 & 20 & 100 & 30 & 100 & 30 & 80 & 75.71\\
\hline
Depot & 0 & 60 & 0 & 90 & 0 & 60 & 40 & 50 & 0 & 30 & 20 & 100 & 10 & 60 & 54.29\\
Goldminer & 0 & 100 & 20 & 100 & 20 & 100 & 20 & 60 & 0 & 90 & 50 & 90 & 10 & 100 & 74.29\\
Satellite & 20 & 60 & 30 & 90 & 10 & 30 & 20 & 50 & 10 & 80 & 0 & 90 & 20 & 100 & 55.71\\
Swap & 20 & 50 & 40 & 40 & 10 & 50 & 30 & 50 & 10 & 80 & 10 & 100 & 40 & 100 & 44.29\\
Alfworld & 30 & 40 & 50 & 90 & 10 & 40 & 20 & 60 & NA & NA & 0 & 20 & 20 & 100 & 36.67\\
\hline
Mean & 15 & 81 & 25 & 89 & 13 & 82 & 34 & 75 & 11 & 87 & 16 & 90 & 20 & 91 & 65.44

\end{tabular}
}
}
\caption{\footnotesize Per-domain comparison of 7 tasks on the multiple choice questions between the Base model, Granite8b code base, and the Finetuned model. The first 8 domains are domains that are in the training set (seen domains), and the last 5 domain are unseen domains. ``Diff'' shows the average difference between the base and fine-tuned model. }
\label{fig:mcqperdomain}
\end{table*}

%% file: all_llms_training.tex
\begin{table*}[t]
    \centering
    \resizebox{\textwidth}{!}{
    \begin{tabular}{l|c@{\hspace{0.6\tabcolsep}}c|c@{\hspace{0.6\tabcolsep}}c|c@{\hspace{0.6\tabcolsep}}c|c@{\hspace{0.6\tabcolsep}}c|c@{\hspace{0.6\tabcolsep}}c|c@{\hspace{0.6\tabcolsep}}c|c@{\hspace{0.6\tabcolsep}}c|c@{\hspace{0.6\tabcolsep}}c}
    \multirow{2}{*}{\textbf{Model}}  & \multicolumn{2}{c|}{Applicability} & \multicolumn{2}{c|}{{Progression}}&  \multicolumn{2}{c|}{Reachability} & \multicolumn{2}{c|}{Validation} & \multicolumn{2}{c|}{Action Reach.} &  \multicolumn{2}{c|}{Justification} & \multicolumn{2}{c|}{Landmark} & \multicolumn{2}{c}{Mean}\\
 & Bool & MCQ & Bool & MCQ& Bool & MCQ& Bool & MCQ& Bool & MCQ& Bool & MCQ& Bool & MCQ& Bool & MCQ\\
 \midrule
Phi-3 128K & $67.50$ & $32.50$ & $71.25$ & $52.50$ & $51.25$ & $30.00$ & $51.25$ & $20.00$ & $51.25$ & $35.00$ & $43.75$ & $32.50$ & $42.50$ & $28.75$ & $54.11$ & $33.04$ \\
Gemma 7B & $64.25$ & $29.25$ & $70.25$ & $34.00$ & $52.50$ & $21.00$ & $50.00$ & $18.75$ & $56.25$ & $39.25$ & $55.00$ & $39.00$ & $23.50$ & $17.75$ & $53.11$ & $28.43$ \\
Granite 7B & $57.50$ & $25.50$ & $63.50$ & $36.25$ & $52.50$ & $32.50$ & $30.00$ & $28.75$ & $51.25$ & $30.00$ & $42.50$ & $25.00$ & $47.50$ & $23.50$ & $49.25$ & $28.79$ \\
Mistral 7B & $55.00$ & $36.25$ & $76.25$ & $37.50$ & $52.50$ & $32.50$ & $49.00$ & $17.50$ & $58.75$ & $21.25$ & $52.50$ & $35.00$ & $42.50$ & $17.50$ & $55.21$ & $28.21$ \\
Mistral instruct 7B & $66.25$ & $32.50$ & $67.50$ & $46.25$ & $62.50$ & $33.75$ & $49.25$ & $35.75$ & $50.00$ & $32.50$ & $37.50$ & $32.50$ & $57.50$ & $42.50$ & $55.79$ & $36.54$ \\
Granite-code 8B & $62.50$ & $36.75$ & $76.25$ & $33.25$ & $52.50$ & $20.75$ & $40.50$ & $20.00$ & $56.25$ & $28.75$ & $50.00$ & $35.00$ & $41.25$ & $18.50$ & $54.18$ & $27.57$ \\
Granite8b code instruct & $53.75$ & $37.50$ & $72.50$ & $34.75$ & $50.00$ & $28.75$ & $49.25$ & $21.25$ & $46.25$ & $44.00$ & $48.75$ & $30.00$ & $40.50$ & $18.75$ & $51.57$ & $30.71$ \\
LLAMA-3 8B & $73.75$ & $52.50$ & $73.75$ & $57.50$ & $56.00$ & $41.75$ & $50.00$ & $45.00$ & $65.00$ & $47.50$ & $55.00$ & $32.50$ & $51.25$ & $27.50$ & $60.68$ & $43.46$ \\
LLAMA-3.1 8B & $61.25$ & $63.75$ & $71.25$ & $52.50$ & $52.50$ & $37.50$ & $62.50$ & $30.00$ & $45.00$ & $31.25$ & $52.50$ & $38.75$ & $31.25$ & $27.50$ & $53.75$ & $40.18$ \\
Mixtral 8x7B & $80.00$ & $56.00$ & $79.25$ & $61.25$ & $77.50$ & $35.00$ & $61.75$ & $36.50$ & $58.50$ & $53.75$ & $51.25$ & $48.75$ & $57.00$ & $46.25$ & $66.46$ & $48.21$ \\
Granite 13B & $42.75$ & $32.50$ & $53.25$ & $22.25$ & $48.75$ & $33.75$ & $50.00$ & $35.00$ & $50.00$ & $23.25$ & $46.25$ & $28.75$ & $53.00$ & $10.00$ & $49.14$ & $26.50$ \\
Codestral 22B & $88.75$ & $41.25$ & $87.50$ & $50.00$ & $55.00$ & $30.00$ & $70.00$ & $27.50$ & $55.00$ & $36.25$ & $67.50$ & $62.50$ & $51.25$ & $25.00$ & $67.86$ & $38.93$ \\
Mixtral 8x22B & $86.00$ & $37.75$ & $77.00$ & $60.00$ & $52.50$ & $43.25$ & $37.50$ & $20.00$ & $56.25$ & $26.50$ & $40.00$ & $44.00$ & $38.25$ & $27.25$ & $55.36$ & $36.96$ \\
Deepseek-33b instruct & $76.25$ & $37.50$ & $73.75$ & $47.75$ & $52.50$ & $34.00$ & $52.50$ & $41.25$ & $52.50$ & $28.75$ & $43.75$ & $30.00$ & $64.00$ & $28.75$ & $59.32$ & $35.43$ \\
CodeLLAMA 34B & $82.50$ & $45.00$ & $78.75$ & $48.75$ & $52.50$ & $24.25$ & $49.00$ & $36.25$ & $43.75$ & $36.25$ & $57.50$ & $37.50$ & $43.75$ & $20.00$ & $58.25$ & $35.43$ \\
\midrule
LLAMA-2 7OB & $78.75$ & $23.75$ & $76.25$ & $40.00$ & $52.50$ & $26.25$ & $48.25$ & $16.25$ & $56.25$ & $20.00$ & $53.75$ & $49.00$ & $23.50$ & $12.50$ & $55.61$ & $26.82$ \\
CodeLLAMA 70B & $77.75$ & $41.50$ & $61.25$ & $55.25$ & $52.50$ & $28.50$ & $40.00$ & $20.00$ & $55.75$ & $27.50$ & $50.00$ & $28.75$ & $37.00$ & $26.25$ & $53.46$ & $32.54$ \\
LLAMA-3 70B & $90.00$ & $86.25$ & $93.75$ & $86.25$ & $87.50$ & $82.50$ & $78.75$ & $52.00$ & $65.75$ & $73.25$ & $61.25$ & $82.50$ & $70.75$ & $60.00$ & $78.25$ & $74.68$ \\
LLAMA-3.1 70B & $95.50$ & $84.25$ & $90.25$ & $90.25$ & $57.75$ & $54.50$ & $64.50$ & $45.50$ & $67.50$ & $62.00$ & $58.75$ & $69.25$ & $29.25$ & $59.50$ & $66.21$ & $66.46$ \\
LLAMA-3.1 405B & $97.50$ & $87.50$ & $93.75$ & $92.50$ & $61.25$ & $80.00$ & $75.50$ & $57.75$ & $70.00$ & $78.75$ & $90.00$ & $82.50$ & $77.50$ & $65.00$ & $80.79$ & $77.71$ \\
\midrule
GPT-4o Mini & $88.75$ & $67.50$ & $96.25$ & $77.50$ & $82.50$ & $37.50$ & $57.50$ & $41.25$ & $55.00$ & $22.50$ & $71.25$ & $63.75$ & $70.00$ & $51.25$ & $74.46$ & $51.61$ \\
GPT-4o & $97.50$ & $90.00$ & $96.25$ & $93.75$ & $78.75$ & $75.00$ & $50.00$ & $45.00$ & $62.50$ & $57.50$ & $83.75$ & $72.50$ & $97.50$ & $71.25$ & $80.89$ & $72.14$ \\

    \end{tabular}
    }
    \caption{Accuracy of \numofmodels{} LLMs on \numoftraining{} seen domains of ACPBench. All models were evaluated with two in-context examples and Chain-of-Thought prompt. The right-most column is mean across tasks.}
    \label{tab:compare_models_train}
\end{table*}

%% file: all_llms_testing.tex
\begin{table*}[t]
    \centering
    \resizebox{\textwidth}{!}{
    \begin{tabular}{l|c@{\hspace{0.6\tabcolsep}}c|c@{\hspace{0.6\tabcolsep}}c|c@{\hspace{0.6\tabcolsep}}c|c@{\hspace{0.6\tabcolsep}}c|c@{\hspace{0.6\tabcolsep}}c|c@{\hspace{0.6\tabcolsep}}c|c@{\hspace{0.6\tabcolsep}}c|c@{\hspace{0.6\tabcolsep}}c}
    \multirow{2}{*}{\textbf{Model}}  & \multicolumn{2}{c|}{Applicability} & \multicolumn{2}{c|}{{Progression}}&  \multicolumn{2}{c|}{Reachability} & \multicolumn{2}{c|}{Validation} & \multicolumn{2}{c|}{Action Reach.} &  \multicolumn{2}{c|}{Justification} & \multicolumn{2}{c|}{Landmark} & \multicolumn{2}{c}{Mean}\\
 & Bool & MCQ & Bool & MCQ& Bool & MCQ& Bool & MCQ& Bool & MCQ& Bool & MCQ& Bool & MCQ& Bool & MCQ\\
 \midrule

Phi-3 128K & $64.00$ & $34.00$ & $64.00$ & $56.00$ & $54.00$ & $20.00$ & $50.00$ & $18.00$ & $57.50$ & $27.50$ & $58.00$ & $36.00$ & $60.00$ & $76.00$ & $58.21$ & $38.21$ \\
Gemma 7B & $61.60$ & $27.60$ & $56.40$ & $26.40$ & $54.00$ & $26.40$ & $42.00$ & $22.00$ & $54.50$ & $25.00$ & $44.00$ & $32.40$ & $34.00$ & $50.40$ & $49.50$ & $30.03$ \\
Granite 7B & $56.00$ & $36.00$ & $42.00$ & $34.00$ & $48.00$ & $38.00$ & $35.60$ & $22.00$ & $42.50$ & $25.00$ & $38.00$ & $26.00$ & $48.00$ & $46.00$ & $44.30$ & $32.43$ \\
Mistral 7B & $72.00$ & $26.00$ & $68.00$ & $40.00$ & $54.00$ & $22.00$ & $46.00$ & $18.00$ & $77.50$ & $15.00$ & $42.00$ & $22.00$ & $24.00$ & $58.00$ & $54.79$ & $28.71$ \\
Mistral instruct 7B & $58.00$ & $30.00$ & $52.00$ & $48.00$ & $60.00$ & $32.00$ & $56.80$ & $36.80$ & $37.50$ & $37.50$ & $52.00$ & $24.00$ & $58.00$ & $64.00$ & $53.47$ & $38.90$ \\
Granite-code 8B & $54.00$ & $25.20$ & $60.00$ & $36.00$ & $52.00$ & $30.00$ & $50.00$ & $12.40$ & $60.00$ & $20.00$ & $42.00$ & $34.00$ & $30.80$ & $62.40$ & $49.83$ & $31.43$ \\
Granite8b code instruct & $58.00$ & $24.00$ & $64.00$ & $34.00$ & $52.00$ & $30.00$ & $40.40$ & $24.00$ & $35.00$ & $30.00$ & $42.00$ & $36.00$ & $49.20$ & $70.00$ & $48.66$ & $35.43$ \\
LLAMA-3 8B & $71.60$ & $44.00$ & $72.00$ & $53.60$ & $54.00$ & $40.00$ & $54.00$ & $56.00$ & $60.50$ & $15.00$ & $61.60$ & $32.00$ & $66.00$ & $70.00$ & $62.81$ & $44.37$ \\
LLAMA-3.1 8B & $72.00$ & $46.00$ & $52.00$ & $40.00$ & $54.00$ & $28.00$ & $56.00$ & $50.00$ & $37.50$ & $22.50$ & $38.00$ & $56.00$ & $38.00$ & $60.00$ & $49.64$ & $43.21$ \\
Mixtral 8x7B & $69.20$ & $60.40$ & $65.60$ & $61.60$ & $73.60$ & $48.00$ & $72.00$ & $32.00$ & $41.50$ & $57.50$ & $62.00$ & $55.60$ & $63.60$ & $82.00$ & $63.93$ & $56.73$ \\
Granite 13B & $40.80$ & $24.00$ & $51.20$ & $18.40$ & $46.00$ & $20.00$ & $54.00$ & $34.00$ & $35.50$ & $32.50$ & $44.00$ & $26.00$ & $46.00$ & $34.00$ & $45.36$ & $26.99$ \\
Codestral 22B & $78.00$ & $36.00$ & $78.00$ & $54.00$ & $54.00$ & $26.00$ & $60.00$ & $20.00$ & $50.00$ & $42.50$ & $68.00$ & $62.00$ & $72.00$ & $70.00$ & $65.71$ & $44.36$ \\
Mixtral 8x22B & $72.40$ & $37.60$ & $64.80$ & $46.00$ & $46.00$ & $41.60$ & $38.00$ & $12.00$ & $63.00$ & $30.50$ & $48.00$ & $45.60$ & $55.20$ & $74.00$ & $55.34$ & $41.04$ \\
Deepseek-33b instruct & $62.00$ & $36.80$ & $60.00$ & $44.00$ & $54.00$ & $28.00$ & $50.00$ & $32.00$ & $45.00$ & $25.00$ & $52.00$ & $20.00$ & $59.60$ & $56.00$ & $54.66$ & $34.54$ \\
CodeLLAMA 34B & $78.00$ & $38.00$ & $64.00$ & $36.00$ & $54.00$ & $28.00$ & $52.00$ & $16.00$ & $72.00$ & $27.50$ & $52.00$ & $32.00$ & $52.00$ & $73.60$ & $60.57$ & $35.87$ \\
\midrule
LLAMA-2 7OB & $78.00$ & $26.00$ & $64.00$ & $31.60$ & $54.00$ & $28.00$ & $56.40$ & $16.00$ & $70.00$ & $26.00$ & $42.00$ & $66.00$ & $26.00$ & $47.60$ & $55.77$ & $34.46$ \\
CodeLLAMA 70B & $70.00$ & $27.60$ & $44.40$ & $49.20$ & $42.40$ & $16.00$ & $40.00$ & $14.00$ & $37.50$ & $31.50$ & $42.00$ & $36.00$ & $37.20$ & $68.00$ & $44.79$ & $34.61$ \\
LLAMA-3 70B & $92.00$ & $76.00$ & $92.00$ & $86.00$ & $88.00$ & $82.00$ & $78.40$ & $64.00$ & $50.00$ & $42.50$ & $64.00$ & $90.00$ & $90.00$ & $72.40$ & $79.20$ & $73.27$ \\
LLAMA-3.1 70B & $89.20$ & $84.40$ & $89.20$ & $81.20$ & $67.20$ & $55.60$ & $68.80$ & $48.40$ & $54.00$ & $50.00$ & $54.00$ & $67.20$ & $43.20$ & $84.80$ & $66.51$ & $67.37$ \\
LLAMA-3.1 405B & $92.00$ & $86.00$ & $92.00$ & $96.00$ & $56.00$ & $82.00$ & $80.00$ & $71.20$ & $55.00$ & $37.50$ & $90.00$ & $94.00$ & $92.00$ & $66.00$ & $79.57$ & $76.10$ \\
\midrule
GPT-4o Mini & $94.00$ & $84.00$ & $94.00$ & $82.00$ & $78.00$ & $42.00$ & $84.00$ & $54.00$ & $52.50$ & $20.00$ & $88.00$ & $80.00$ & $88.00$ & $94.00$ & $82.64$ & $65.14$ \\
GPT-4o & $96.00$ & $88.00$ & $92.00$ & $84.00$ & $80.00$ & $80.00$ & $80.00$ & $68.00$ & $47.50$ & $42.50$ & $96.00$ & $94.00$ & $92.00$ & $92.00$ & $83.36$ & $78.36$ \\

    \end{tabular}
    }
    \caption{Accuracy of \numofmodels{} LLMs on \numoftesting{} unseen domains of ACPBench. All models were evaluated with two in-context examples and Chain-of-Thought prompt. The right-most column is mean across tasks.}
    \label{tab:compare_models_test}
\end{table*}